\documentclass{article} % For LaTeX2e
\usepackage[preprint]{colm2025_conference}
\usepackage{xspace}
\usepackage{wrapfig}
\usepackage{microtype}
\usepackage{hyperref}
\usepackage{url}
\usepackage{booktabs}
\usepackage{amssymb}
\usepackage{lineno}
\usepackage{amsmath}
\usepackage[capitalise]{cleveref}
\usepackage{tikz}
\usepackage{pifont}
\usetikzlibrary{positioning, shapes, arrows}
\usepackage{xcolor}
\definecolor{darkblue}{rgb}{0, 0, 0.5}
\hypersetup{colorlinks=true, citecolor=darkblue, linkcolor=darkblue, urlcolor=darkblue}
\usepackage{tipa}
\usepackage{longtable}
\usepackage{booktabs}  % For better table formatting (optional)

\title{FineWeb2: One Pipeline to Scale Them All — Adapting \\Pre-Training Data Processing to Every Language}

\newcommand{\huggingface}{\raisebox{3pt}{\includegraphics[height=0.7em]{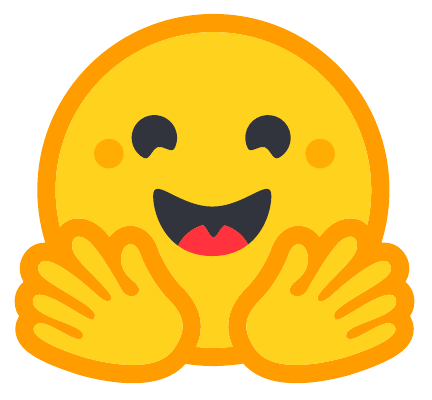}}}
\newcommand{\epfl}{\raisebox{3pt}{\includegraphics[height=0.4em]{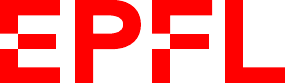}}}

\newcommand{\huggingfacedown}{\includegraphics[height=0.75em]{assets/hf-logo.pdf}\xspace}
\newcommand{\githubdown}{\includegraphics[height=0.75em]{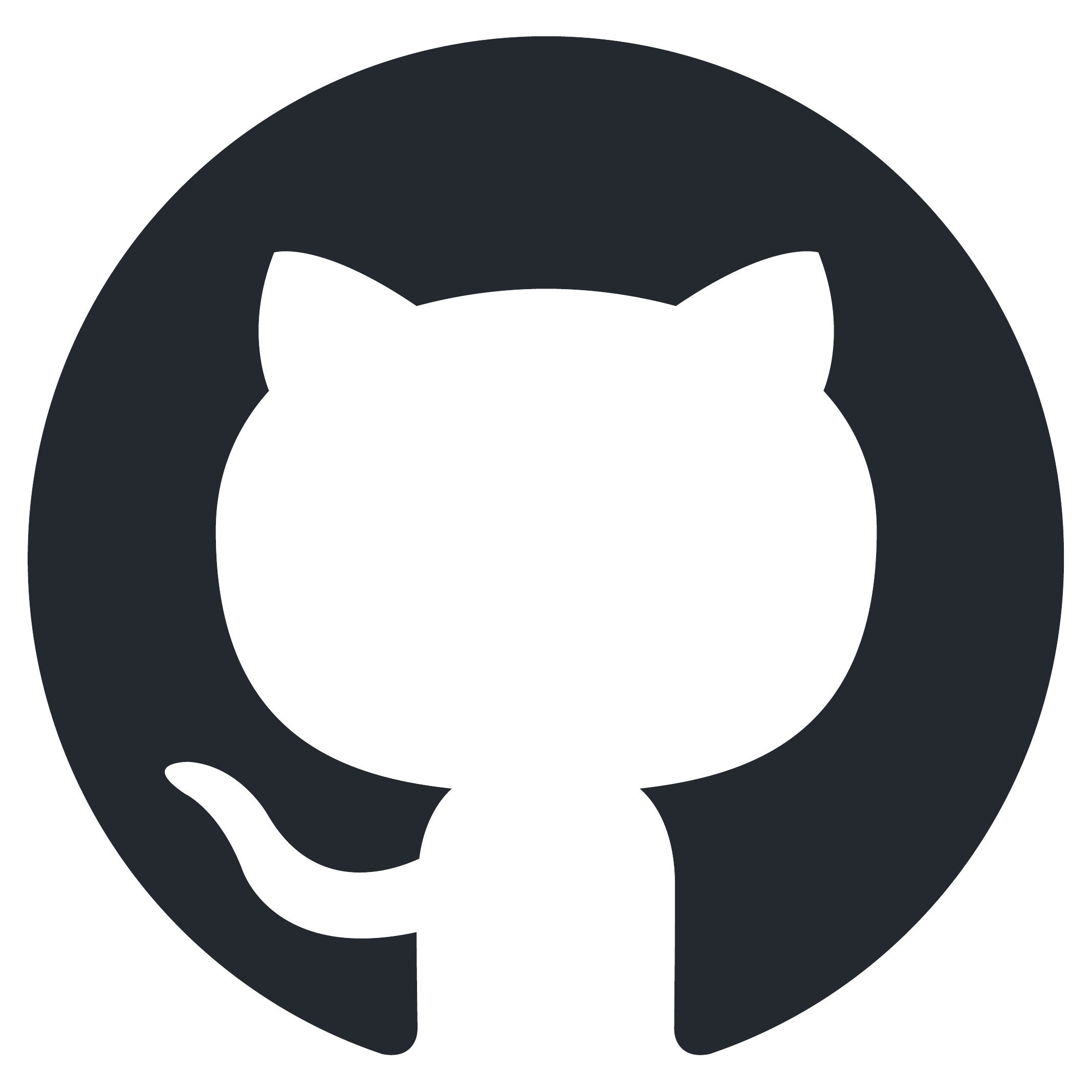}\xspace}

\author{Guilherme Penedo\huggingface\thanks{Correspondence to guilherme at huggingface dot co} \quad Hynek Kydlíček\huggingface \quad Vinko Sabolčec\epfl\quad Bettina Messmer\epfl\\ \textbf{Negar Foroutan}\epfl\quad \textbf{Amir Hossein Kargaran}\huggingface \quad \textbf{Colin Raffel}\huggingface \quad \textbf{Martin Jaggi}\epfl\\ \textbf{Leandro Von Werra}\huggingface \quad \textbf{Thomas Wolf}\huggingface
 \\
  \centerline{\huggingface Hugging Face \quad \epfl EPFL}
  \\\\
  \quad \quad \quad \githubdown Pipeline code: \url{https://github.com/huggingface/fineweb-2}\\
  \quad \quad \quad \huggingfacedown FineWeb2 dataset: \url{https://hf.co/datasets/HuggingFaceFW/fineweb-2}
\vspace{-2em}
}
\begin{document}

\ifcolmsubmission
\linenumbers
\fi

\maketitle

\begin{abstract}
Pre-training state-of-the-art large language models (LLMs) requires vast amounts of clean and diverse text data. While the open development of large high-quality English pre-training datasets has seen substantial recent progress, training performant multilingual LLMs remains a challenge, in large part due to the inherent difficulty of tailoring filtering and deduplication pipelines to a large number of languages. In this work, we introduce a new pre-training dataset curation pipeline based on FineWeb \citep{penedoFineWebDatasetsDecanting2024} that can be automatically adapted to support any language. We extensively ablate our pipeline design choices on a set of nine diverse languages, guided by a set of meaningful and informative evaluation tasks that were chosen through a novel selection process based on measurable criteria. Ultimately, we show that our pipeline can be used to create non-English corpora that produce more performant models than prior datasets. We additionally introduce a straightforward and principled approach to rebalance datasets that takes into consideration both duplication count and quality, providing an additional performance uplift. Finally, we scale our pipeline to over 1000 languages using almost 100 Common Crawl snapshots to produce FineWeb2, a new 20 terabyte (5 billion document) multilingual dataset which we release along with our pipeline, training, and evaluation codebases.
\end{abstract}

\section{Introduction}
% I don't absolutely love it here but it gives more space overall
\begin{wrapfigure}{r}{0.5\linewidth}
    % \vspace*{-1.2\baselineskip}  % pull the image up to align with first text line
    \vspace*{-4\baselineskip}  % pull the image up to align with intro heading
    \centering
    \includegraphics[width=\linewidth]{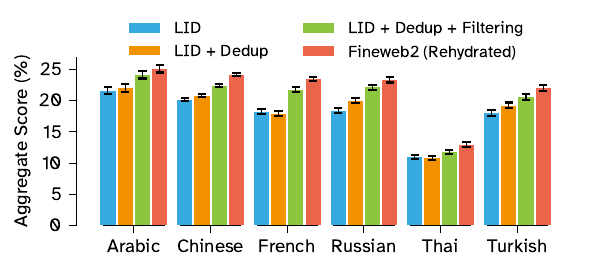}
    \vspace{-1\baselineskip}
    \caption{\textbf{The FineWeb2 pipeline:} Evaluation results of models trained on 350 billion tokens show that each pipeline step -- Language Identification (\textbf{LID}), Deduplication (\textbf{Dedup}), \textbf{Filtering}, and Dedup-informed upsampling (\textbf{Rehydration}) -- improves performance.
    % Interestingly, each step has a different impact per language.
    }
    \label{fig:all_stages_impact}
\end{wrapfigure}
One of the main drivers of the improving capabilities of large language models (LLMs) is increased scale, in terms of both model and pre-training dataset size.
To satiate the ever-growing hunger for text data, most pre-training datasets include large amounts of text scraped from the public internet~\citep{raffel2020exploring,penedo2023refinedwebdatasetfalconllm,penedoFineWebDatasetsDecanting2024}.
Consequently, pre-training data tends to be most readily available in the ``high-resource'' languages (English, Chinese, etc.) that are most prevalent on the internet.
Since LLM capabilities largely stem from the data they were trained on~\citep{grosse2023studying,roberts2020much,razeghi2022impact}, this has resulted in language models having better performance on high-resource languages.
Furthermore, commercial and open language model development frequently only targets these languages~\citep{grattafiori2024llama3herdmodels,jiang2024mixtralexperts,ai2025yiopenfoundationmodels}.
This state of affairs leaves the majority of the world's population (speaking over 7,000 languages~\citep{ethnologue}) unable to interact with state-of-the-art LLMs in their native tongue.

Why not just curate datasets in underrepresented languages and train LLMs on them?
Putting aside the possible lack of data (recent LLM training runs typically require trillions of tokens~\citep{llama3modelcard,deepseekai2024deepseekllmscalingopensource}), a key challenge is that high-resource languages benefit from the existence of well-tuned and battle-tested data processing and curation pipelines, whereas low-resource languages face a vastly different landscape:
evaluating corpora quality, ensuring accurate language identification, customizing filtering recipes, and even separating words can be major challenges for many languages.
While some past work has successfully curated single-language pre-training datasets and used them to produce strong language-specific models~\citep[etc.]{carmo2020ptt5,de2019bertje,le2019flaubert,martin2019camembert,delobelle2020robbert,luukkonen2023fingpt,pllum2025,pipatanakul2023typhoon}, hand-designing a different pipeline for each language does not scale.
Consequently, most past work on multilingual datasets (e.g.~\citealp{xue2021mt5massivelymultilingualpretrained,wenzek-etal-2020-ccnet,de2024new}) has used a (mostly) fixed pipeline across all languages.
This one-size-fits-all approach risks applying inappropriate filtering to different languages, obviating the goal of creating performant data in many languages.

In this work, we introduce a new data processing pipeline based on the approach used for the state-of-the-art English pre-training dataset FineWeb~\citep{penedoFineWebDatasetsDecanting2024}.
Importantly, our pipeline can be \textit{automatically adapted} based on language-specific statistics to produce high-quality pre-training corpora in any language. 
We follow a data-driven approach and validate our design choices by running extensive ablation experiments where we train monolingual models on a set of nine diverse languages and evaluate on tasks chosen through a novel selection process based on measurable criteria that ensure a meaningful signal. 
% don't love the transition here
In addition, we introduce a straightforward and principled approach to rebalance datasets using the original duplication counts and quality signals that allows globally near-deduplicated datasets to obtain a performance uplift.
Ultimately, we show that models trained on language-specific corpora produced by our pipeline perform better than those trained on other public web-based multilingual datasets by training models on additional ``unseen'' languages that were not used to inform pipeline design decisions.
Finally, we use our pipeline to process almost 100 Common Crawl\footnote{\url{https://www.commoncrawl.org/}} snapshots spanning the summer of 2013 to April 2024 to create FineWeb2, a new 20 terabyte (5 billion document) dataset covering over 1000 languages.
FineWeb2 is released under the permissive ODC-By License, and we additionally release the pipeline, training, and evaluation codebases, as well as the preliminary version of the dataset obtained after the deduplication stage, to facilitate further research on multilingual pre-training dataset curation.
% Overall, our work provides significant resources to facilitate research and development of LLMs on a large number of languages.
% \section{Background}

\hypertarget{preliminaries}{%
\section{Preliminaries}\label{preliminaries}}

Before detailing our dataset creation process, we first establish critical considerations that arise when dealing with massively multilingual data.

\paragraph{Notation}
When considering thousands of languages, it's important to have an
unambiguous way of referring to languages and scripts. In our work we
identify languages by their official ISO-639-3 codes\footnote{\url{https://iso639-3.sil.org/code_tables/639/data}} which cover
significantly more languages than the commonly used ISO-639-1 codes
(such as ''en'', ''zh'', etc).
As many languages use multiple writing systems (scripts), we optionally
designate individual ''languages'' by a \textit{(ISO-639-3 language code, ISO 15924 script code)} pair. For
instance, `arb\_Arab` is Standard Arabic in Arabic script, while
`arb\_Latn` is Standard Arabic in Latin script.

\paragraph{Separating words}
\label{sec:separating-words}
Many parts of our processing and evaluation pipeline require the ability
to separate (tokenize) text into individual words. For example, we rely on word tokenization
when we filter documents based on the ratio of words that have a given
property, when selecting n-grams for deduplication, or even when
evaluating generative tasks.
While whitespace and punctuation often mark word boundaries, many
writing systems use different boundary markers or have no visible
markers at
all~\citep{daniels1996worlds}. This is particularly common in Southeast Asian languages, as well as
Chinese, Japanese, and Korean.
Therefore, word tokenizers/segmentators tailored to each language and script are
needed. We collected a large number of tokenizers from SpaCy~\citep{Honnibal_spaCy_Industrial-strength_Natural_2020} and Stanza~\citep{qi2020stanzapythonnaturallanguage}, as well as from libraries targeting specific languages (or language groups). We then assigned proxy tokenizers based on the closest language according to language family data from the
Ethnologue\footnote{\url{https://www.ethnologue.com/browse/families/}} to languages without a native word tokenizer. For more details on this process, see~\cref{app:word-tokenizers}. These tokenizer assignments were crucial to adapt filtering, deduplication, and evaluation setups to thousands of languages.

\hypertarget{experimental-setup}{%
\section{Experimental setup}\label{experimental-setup}}

To compare and validate pipeline design choices, we followed an experimental setup similar to~\citet{penedoFineWebDatasetsDecanting2024}. Specifically, to assess
data quality, we relied on training small models and evaluating them on
``early-signal'' benchmark tasks, i.e., tasks where models perform reasonably well after only a few tens of billions or hundreds of billions of training tokens, allowing us to confidently establish comparisons between them. For each processing step, we conducted
comparative evaluations using two identical models that differed only in
their training data: one model was trained on data with the processing
step applied, while the other used the unprocessed (ablated) version. By
keeping all other variables constant (number of parameters,
architecture, tokenizer, and training token count), we could isolate the
impact of each data processing step on downstream model quality.

While ideally we would have tested each processing step across every language, computational constraints and the lack of
evaluation tasks for many of the languages made this impractical. We
therefore chose to conduct our experiments on a select set of nine \textbf{canary languages} (i.e.\ test languages): Arabic, Chinese, French, Hindi, Russian, Swahili,
Telugu, Thai, and Turkish. Testing across these languages allowed us to
evaluate the impact of each design decision across different language
families, scripts, and levels of resource availability, while keeping computational requirements manageable. These details are available on~\cref{tab:canary-languages}, where Resource Availability was determined following~\citet{joshi-etal-2020-state}. We trained separate
models per language, rather than a single multilingual model, to avoid
introducing confounders between languages. This means that for every
ablation experiment or validation run reported in this paper, \emph{we trained nine different models (one per language)}.

\hypertarget{tokenizer-and-model-architecture}{%
\subsection{Tokenizer and model architecture}\label{tokenizer-and-model-architecture}}

\paragraph{Tokenizer}
The choice of tokenizer can induce differential downstream model performance across different languages based on how compactly it maps a given language's words into tokens~\citep{mielke2021between}.
Given that our experiments target different languages and, in particular, different scripts, we evaluated the \textbf{subword fertility} and \textbf{proportion of continued words}~\citep{rust2021goodtokenizermonolingualperformance} of different existing open-source tokenizers from leading multilingual LLMs on our nine canary languages. Concretely, we split text from each language's Wikipedia into individual ''real'' words using our word-level tokenizers (discussed in \cref{app:word-tokenizers}) and then measured the average number of tokens per word for each tokenizer. From the tokenizers that showed reasonable fertility on our nine canary languages, we chose the tokenizer used in \textbf{Gemma}~\citep{gemmateam2024gemmaopenmodelsbased}, a modern tokenizer with a vocabulary size of around 250,000 tokens that showed better average fertility than similarly sized tokenizers. Detailed results 
% and comparisons
are available in~\cref{app:tokenizer-metrics}. % To simplify our setup, we use the Gemma tokenizer for all languages for which we train monolingual models.

\paragraph{Model architecture}
We used a similar model architecture setup to~\citet{penedoFineWebDatasetsDecanting2024}, with a reduced number of layers given the additional embedding parameters due to the larger vocabulary size. All models used in our experiments were trained using the \href{https://github.com/huggingface/nanotron}{nanotron} training framework, and followed the Llama~\citep{touvron2023llamaopenefficientfoundation} architecture with 14 layers, 32 attention heads, length-2048 sequences, and tied embeddings, for a total of \textbf{1.46 billion}
parameters. Further details and training hyperparameters are provided in~\cref{app:model-architecture}.

\subsection{Baseline datasets}
\label{sec:reference-datasets}

We selected existing widely used multilingual datasets to use as comparison baselines. For each language, we trained one model on language-specific data from each reference dataset: \textbf{CC-100}~\citep{wenzek-etal-2020-ccnet,conneau-etal-2020-unsupervised}, \textbf{mC4}~\citep{xue2021mt5massivelymultilingualpretrained}, \textbf{CulturaX}~\citep{nguyen-etal-2024-culturax}, and \textbf{HPLT}~\citep{degibert2024newmassivemultilingualdataset}. We additionally trained multiple models on ''raw'' Common Crawl data (after text extraction and Language Identification, but without any additional filtering or deduplication).
Unfortunately, all datasets except raw Common Crawl only contained a limited amount of data for Telugu and Swahili, and only CulturaX and HPLT had
enough data for a pre-training run in Hindi at 30 billion tokens without requiring an excessive number of epochs over the training data.

\hypertarget{selecting-evaluation-finetasks}{%
\subsection{Selecting evaluation
(Fine)tasks}\label{selecting-evaluation-finetasks}}

The selection of English evaluation tasks is straightforward due to the
existence of well-established benchmarks such as MMLU~\citep{hendryckstest2021} or
HellaSwag~\citep{zellers2019hellaswagmachinereallyfinish}, which are widely used and supported by all
major evaluation frameworks. The situation is significantly different
for non-English languages, which often lack evaluation tasks. When available, these tasks often lack broader community validation and suffer from quality
issues -- many are machine-translated and may even include English words
in their formulations~\citep{Artetxe_2020}. Additionally, we find that
non-English tasks are often unsuitable for early pre-training evaluation
due to suboptimal task formulations and/or excessive difficulty that
results in random-level performance.

To identify informative evaluation tasks, we established four key
criteria for what we call \textbf{early-signal} tasks: \textbf{Monotonicity} -- the performance of models evaluated on this
  task should improve as training progresses, though possibly at different rates depending on the pre-training dataset; \textbf{Low noise} -- when comparing models trained on different
  datasets, we want to ensure that the relative performance differences
  between them are due to inherently better training data, and not due
  to evaluation noise; \textbf{Non-random performance early in training} -- tasks reflecting model capabilities that are only acquired later in training are not
  informative for small scale pre-training ablations, as near-random
  scores cannot meaningfully differentiate between datasets; \textbf{Ordering consistency} -- if model A outperforms model B, then
  falls behind, then leads again within a short span of training steps,
  we cannot confidently determine which model (and, correspondingly,
  dataset variant) is superior and we therefore need tasks that provide consistent relative performance.

We defined quantitative metrics to measure these characteristics and applied them to hundreds of candidate zero-shot evaluation
tasks targeting our 9 canary languages on the models trained on our baseline datasets.  See~\cref{app:Fine-Tasks} for the precise definition of ``early-signal'' tasks and additional description of our evaluation setup. We strove to cover
\textbf{different task types} in all languages: Reading Comprehension,
\textbf{RC}; General Knowledge, \textbf{GK}; Natural Language
Understanding, \textbf{NLU}; and Common-Sense Reasoning, \textbf{CR}. 

Our in-depth analysis of existing evaluation tasks resulted
in a final suite of \textbf{84} selected benchmarks out of \textbf{197} tested across our nine canary
languages. We list all the tasks and employed metrics in \cref{app:finetasks-selection}.
% We ran all evaluations in a 0-shot setting.
%

% \hypertarget{aggregating-scores}{%
% \subsection{Aggregating scores}\label{aggregating-scores}}

To produce an \textbf{aggregate score} across tasks, we follow the approach used by~\citet{open-llm-leaderboard-v2, li2024datacomplmsearchgenerationtraining} and average scores across tasks after first rescaling scores based on the random baseline -- any score below the random baseline is considered 0, and for the remaining scores we subtract the random baseline value and shift the scores as $new\_score = (score - random\_baseline) / (1 - random\_baseline)$.
% there would ideally be a single number combining
% the scores from individual tasks -- an \textbf{aggregate score}. A
% simple average of scores would make tasks with fewer options (e.g.~a
% task with 2 options and therefore 50\% random baseline score) have a
% greater impact on the final score than tasks such as a 4 option task
% (and thus with 25\% random baseline score). Additionally, the different metrics we use (F1 score and normalized probabilities) have widely different scales which further conflates reliable averaging.
As some languages might have an unbalanced number of tasks for each task
category (RC, GK, NLU and CR), during score averaging we first average
within categories themselves and then take the average of each
category. This per-category macro-average score is our final reported
aggregate score.
% trying to save some space
%for each language.

\hypertarget{building-fineweb2}{%
\section{The FineWeb2 pipeline}\label{building-fineweb2}}

\hypertarget{starting-point-fineweb}{%
\subsection{Starting point: FineWeb}\label{starting-point-fineweb}}

We started by applying the first few processing steps used in the creation of the English-only FineWeb dataset~\citep{penedoFineWebDatasetsDecanting2024}: downloaded WARC (web archive) files from all available (almost 100) CommonCrawl snapshots, applied URL filtering using a
  \href{https://dsi.ut-capitole.fr/blacklists/}{blocklist} to remove
  adult content (an approach discussed in \citet{penedo2023refinedwebdatasetfalconllm}), and used trafilatura~\citep{barbaresi-2021-trafilatura} to extract text content from the HTML in the WARC
  files.
We then aimed to adapt the remaining components of the FineWeb pipeline --
filtering and deduplication -- starting with all the data
that was excluded during FineWeb's language filtering step (which uses the FastText language identifier~\citep{joulin2016fasttextzipcompressingtextclassification} to identify English text with a threshold of 0.65).
Since approximately 40\% of all documents met the FineWeb English language
threshold, our starting point for FineWeb2 comprises the remaining 60\%
of all the text extracted from CommonCrawl content.

% \begin{wrapfigure}{r}{0.5\linewidth}
%     \centering
%     \includegraphics[width=\linewidth]{plots/all_stages_improvement_small.pdf}
%     \caption{\textbf{The FineWeb2 pipeline:} Evaluation results of models trained on 350 billion tokens show that each pipeline step -- Language Identification (\textbf{LID}), Deduplication (\textbf{Dedup}), \textbf{Filtering}, and Dedup informed upsampling (\textbf{Rehydration}) -- each provide a performance uplift. Interestingly, each step has a different impact per language.}
%     \label{fig:all_stages_impact}
% \end{wrapfigure}

\hypertarget{language-identification-lid}{%
\subsection{Language Identification
(LID)}\label{language-identification-lid}}

A critical first step for curating a multilingual dataset from web scrapes is accurately identifying the main language of each document. The choice of
Language Identification (LID) tool determines not only how reliably each
language (label) is predicted, but also the set of identifiable languages -- if the LID does not have a label for a specific
language, then its content will either be removed or misclassified as
some other language. Additionally, as LID classifiers usually assign a
confidence score to each prediction, the choice of filtering thresholds
further affects the amount of data retained, as well as its quality,
as LID confidence can often be correlated with the noisiness of a given document~\citep{nllbteam2022languageleftbehindscaling}.

% \hypertarget{choosing-a-classifier}{%
% \subsubsection{Choosing a classifier}\label{choosing-a-classifier}}

\paragraph{Choice of classifier} While Transformer-based LID classifiers exist~\citep{bapna2022buildingmachinetranslationsystems}, they are too slow
and expensive to run at a large scale. Most commonly used LID classifiers
are simple character level n-gram models, including CLD3~\citep{salcianu2018compact} (107 supported languages, 
used in mC4~\citep{xue2021mt5massivelymultilingualpretrained}) and classifiers following the fastText architecture~\citep{joulin2016fasttextzipcompressingtextclassification}, such as FT176\footnote{\url{https://fasttext.cc/docs/en/language-identification.html}}
(176 languages, used in CC-100~\citep{wenzek-etal-2020-ccnet,conneau-etal-2020-unsupervised} and CulturaX~\citep{nguyen-etal-2024-culturax}), OpenLID~\citep{burchell-etal-2023-open} (193
languages, used in HPLT2~\citep{burchell2025expandedmassivemultilingualdataset}), and the recent GlotLID~\citep{kargaran-etal-2023-glotlid} (1880 languages).
Although FineWeb~\cite{penedoFineWebDatasetsDecanting2024} used FT176, using GlotLID would allow us to support a much larger number of languages, as well as to run separate processing for different scripts of the same language, as GlotLID explicitly separates them. Additionally, it includes special labels for non supported scripts and for common formats of ``noise'' documents, preventing this content from being classified as one of the other languages.

While GlotLID reports strong performance on language classification
benchmarks and supports a large number of languages, we are primarily interested in the downstream model quality
resulting from using a given LID tool. Therefore, for each canary
language we trained one model on documents classified as this language
(regardless of confidence) by FT176 and another based on GlotLID. We
then evaluated the models on our set of evaluation tasks and found that GlotLID outperforms FT176 (\cref{fig:ft176vsglotlid}) on higher resource languages while being slightly behind on lower resource languages. We consider the increased language coverage to make up for this difference and employ GlotLID for our pipeline. See~\cref{app:classifier-choice} for additional discussion and results.

% \hypertarget{confidence-thresholds}{%
% \subsubsection{Confidence thresholds}\label{confidence-thresholds}}

\paragraph{Confidence thresholds} In addition to providing the most likely language of a document, LID
classifiers typically also return a confidence threshold for that
prediction. Many works rely on a single confidence
threshold applied to all languages, e.g., in mC4~\citep{xue2021mt5massivelymultilingualpretrained} only documents whose
language prediction score is above 70\% are kept, while in CC-100~\citep{wenzek-etal-2020-ccnet} a score of 50\% is used for all languages. However, this does not
account for inherent differences in prediction confidence between
languages -- some languages have a closely related cousin that might
confound the LID classifier, therefore requiring a lower threshold,
whereas a higher value can be employed for high resource languages for
which the classifier is often quite confident~\citep{nllbteam2022languageleftbehindscaling}. To determine appropriate thresholds per language following
our data-driven philosophy, we train models for each of our nine languages
at different confidence thresholds, corresponding to removal rates of
5\% of the data at a time.

Languages such as Arabic (\cref{tab:lang_threshold_ar}) or Russian (\cref{tab:lang_threshold_ru}) prefer high thresholds (\textgreater0.8), while for Swahili a lower threshold around 0.3 (corresponding to a removal rate of almost 65\%) performs best, as this language's distribution is right-skewed. After analyzing the score distributions and the highest performing thresholds, we defined filtering thresholds to be one standard deviation below the median of the score distributions, clipped to the range $\left[0.3, 0.9\right]$: $\max\{0.3, \min\{0.9, \text{Med}(X) - \sigma(X)\}\}$, where $X$ is the distribution of confidence scores for this language's data. We found that this formula selects values within the highest performing threshold regions for most languages (\cref{tab:language_threshold_parameters}).
%median - std\_dev of the score distributions, clipped to {[}0.3, 0.9{]}:\\ `max(0.3, min(0.9, median - std\_dev))`

\hypertarget{deduplication}{%
\subsection{Deduplication}\label{deduplication}}

Deduplication is the process of removing highly similar documents
from a pre-training dataset to increase training efficiency and improve model
performance~\citep{lee2022deduplicatingtrainingdatamakes}. While deduplication requires a large amount of computation and is therefore typically applied as the very last processing step, we employ it as an initial step, before filtering. This allowed us to directly observe the final dataset performance each time we ran one of our many filtering experiments without the possibility of deduplication later influencing the results.

% The FineWeb pipeline {[}link v1{]} established that the scale at which
% deduplication is performed can have a large impact on dataset quality:
% deduplicating individual CommonCrawl snapshots produced better models
% than deduplicating the entire dataset as a whole. However, per-snapshot
% resulted in a significantly larger dataset that is more expensive to
% store. As a compromise, 
We rely on MinHash~\citep{broder1997resemblance}, a ``fuzzy'' deduplication method that
finds clusters of similar documents that are then filtered to keep a
single document per cluster.
We used the same MinHash hyperparameters used for FineWeb (14 buckets of size 8, with 5-grams) and deduplicated globally per language. We used our word-level tokenizers (\cref{sec:separating-words}) to obtain word n-grams.
When keeping a single document per duplicate
cluster, we record the number of documents that were in the cluster to explore duplication-aware upsampling schemes later in \cref{rehydration}.

To measure the impact of deduplication on data quality, we trained per-canary-language
models on 350 billion tokens, both on the data before deduplication
(with the LID filtering) and after. Results in~\cref{fig:all_stages_impact} show that while we generally observed improved performance across languages, the
impact of deduplication seems to vary significantly from language to
language, without any discernable relationship to the language's resource level.
However, we note that even languages showing little to no improvement from deduplication still benefit from rehydration (our duplication-aware upsampling scheme, described in~\cref{rehydration}).
% We leave exploration of the disparate per-language impact of deduplication for future work.
% while in Russian and Turkish we observe a large performance
% uplift, in Arabic and French this difference is much smaller, and for
% Chinese and Thai there does not seem to be any observable improvement.
% This difference does not seem to be due to a language's corpus size,
% given that Chinese and Russian (where deduplication had a small and
% large impact respectively) are our highest resource languages.
% Additionally, we explored whether varying the n-gram size (the number of
% consecutive words) used to compute MinHash signatures could play a role,
% since Chinese ``words'' are usually made up of fewer characters on
% average, but could not find a meaningful difference when trying 3-, 4-,
% 6-, 7-, 8-, 9- or 10-grams. We leave further exploration for future
% work.
% We apply our global MinHash deduplication pipeline to every
% language that still retained data after undergoing the Language Score
% threshold filtering.

\hypertarget{filtering-recipe}{%
\subsection{Filtering recipe}\label{filtering-recipe}}

Filtering aims to remove documents
that are deemed to be ``lower-quality'' (i.e.\ those that might worsen model
performance) using heuristic rules,
such as the number of times words are repeated within the document, the average number of
characters per words in the document, or the ratio of lines ending with punctuation~\cite{albalak2024survey}.
Unfortunately, many of these rules are language-specific: in languages like
Chinese, words have, on average, fewer characters, while in languages
like German the opposite is true.

We began with the list of filtering rules from FineWeb and sought to devise
methods that would allow us to automatically adapt them to a large number of
languages, tailoring specific thresholds according to each language's
characteristics. To this end, we collected statistics for each language on
different corpora and used the distributions on different metrics to
determine adequate filtering thresholds. We relied on our nine canary
languages to inform our decisions and trained a large number of models
to test how well each rule adaptation method would generalize.
% \hypertarget{reference-data}{%
% \subsubsection{Reference data}\label{reference-data}}
We leveraged three main sources to collect statistics for each language:
Wikipedia, the Glotlid-Corpus~\citep{kargaran-etal-2023-glotlid} (used
to train the GlotLID classifier) and our language-filtered data obtained
from Common Crawl.

\hypertarget{stopwords}{%
\subsubsection{Stopwords}\label{stopwords}}

Stopwords are common words in a language that, while not indicative of
text quality, when absent can help identify non-linguistic
''low-quality'' content (e.g.~boilerplate, non-natural text, or
gibberish), or content whose language was misclassified. The number of
stop words in a document is therefore used as a signal to remove such
data, and stopword filtering is part of the widely used Gopher quality
filters~\citep{rae2022scalinglanguagemodelsmethods} for English.

To determine stopwords for each language, we analyzed word frequencies
in our reference datasets, using our word tokenizers to
identify the most frequently occurring words. Instead of selecting a
fixed number of words, we defined stop words as those exceeding a set frequency
threshold. This method allowed us to account for variations across
languages. For example, in English, ``the'' is highly frequent, whereas
in German, its equivalents---``der,'' ``die,'' and ``das''---share the
same role. We additionally addressed specific issues: some ``words'' were actually non-alphabetic and had to be excluded, and for some languages the source data (particularly Wikipedia) contained large portions of English content that caused a significant number of the stop words to be in English. This underscores the importance of having clean data when creating
filters in an automated fashion. Further discussion in~\cref{app:stopwords}. For our filtering pipeline, we require at least 2 words from the stopwords list to be present in each document, in line with~\citet{rae2022scalinglanguagemodelsmethods}.

\hypertarget{filtering-threshold-selection}{%
\subsubsection{Filtering threshold
selection}\label{filtering-threshold-selection}}

To automatically determine filter thresholds for different languages, we propose an empirical approach based on the distribution of the metric we are filtering. %on in each language. 
We consider a variety of different methods: \textbf{English}, use English-based filtering values from FineWeb without change (one of the baselines); \textbf{MeanStd}, assuming the threshold is $n$ standard deviations
  from the mean in the metric distribution in English, we set the threshold to the
  corresponding distance from the mean in the target language
  distribution (a variation using the median instead of mean produces similar values); \textbf{Quantile}, where we define the threshold for each language so as to
  remove the same fraction of data as the English threshold removes in
  English; \textbf{10Tail}, inspired by CulturaX~\citep{nguyen-etal-2024-culturax}, we select a threshold
  to remove the `tail' -- exactly 10\% -- of the reference data; \textbf{MedianRatio}, inspired by HPLT2~\citep{degibert2024newmassivemultilingualdataset}, thresholds are selected such that the ratio between English and the
  target language matches the ratio of the medians of English and the
  target language on this metric. For each method, thresholds are computed on different reference corpora for each filter and then models are trained on the data filtered using these filters. We then compare method for each filter across all languages with each other, as well as with a ``no filtering'' baseline.

Precisely, we computed thresholds for each filter used in three of the FineWeb filter groups:  Gopher Quality (goq), Gopher Repetition (gor), and FineWeb Quality (fwq). We then trained nine models (one per canary language) on data filtered using each method on each of the filter groups, for all method-filter group combinations except those that removed an excessive amount of data (more than 75\%), or that did not remove any data at all. In total, these experiments required a total of 207 ablation models, each trained for 29B tokens. We report the average rank of the aggregate score of each method across languages, in~\cref{tab:filtering-runs-table}. Ultimately, we employ the best performing methods for each filter group: the \textbf{10Tail} method and \textbf{Quantile} methods computed on Wikipedia (or on GlotLID-Corpus for languages without a Wikipedia) for the FineWeb and Gopher Quality filters, respectively, and the \textbf{MeanStd} method computed on Common Crawl data for the Gopher Repetition filters. This step noticeably improves performance for all languages (\cref{fig:all_stages_impact}).

\subsubsection{Precision filtering lower-resource languages}

Low-resource languages often suffer from low LID precision: due to the large class imbalance between high- and low-resource languages on web corpora, real precision is often much lower than that measured on a balanced test set~\citep{caswell-etal-2020-language}. In practice, this means that corpora for low-resource languages with a closely related high-resource language are often heavily contaminated with false positives from the high-resource language, sometimes accounting for more than 90\% of the data.

After inspecting data for low-resource languages produced by our pipeline, we decided to employ a final filtering step exclusively to low-resource languages to address this issue. Inspired by~\citet{caswell-etal-2020-language, bapna2022buildingmachinetranslationsystems}, we compiled lists of words that are common in each language but uncommon in other languages (i.e., have high affinity). We then measured the ``contamination'' of each corpora as the ratio of documents not containing any of these words. While the majority of languages had extremely low contamination scores, roughly a third of the 1900 languages had contamination scores above 10\%. For these languages, we filtered documents using the high-affinity wordlists to remove false positive documents. Additionally, since we noticed the high-affinity wordlists could be too short and strict for some languages (such as English-based pidgins, for example), we also kept documents removed by the wordlist filtering whose URLs included specific terms related to the language (the language code, the language name, domain name extensions etc). A manual audit of three lower-resouce languages shows precision improvements of almost 30\% for some languages. We provide additional details in~\cref{app:precision-lower-resource}.
% TODO mention the audit results

% On our medium scale ablations, filtering improves performance across the board (\cref{fig:all_stages_impact}).

\hypertarget{rehydration}{%
\subsection{Rehydration}\label{rehydration}}

In contrast to standard deduplication practices~\citep{lee2022deduplicatingtrainingdatamakes}, ~\citet{penedoFineWebDatasetsDecanting2024} makes the case for per-snapshot deduplication and claims that additional deduplication beyond the removal of the largest duplicate clusters may actually harm model performance by artificially upsampling documents that are completely unique but high-entropy and low-quality.
While we perform global deduplication, as mentioned in~\cref{deduplication}, we also save the original size of each duplicate cluster in the metadata of the kept documents, which allows us to selectively upsample specific documents (and
therefore ``rehydrate'' the dataset), to obtain more performant models.

In~\citet{txt360data}, the authors explore one such strategy with hand-picked
upsampling weights based on MinHash cluster sizes: documents
with 2 to 5 duplicates are repeated 3 times, 5-100 5 times, 101-1000 8
times, and documents with over 1000 duplicates are repeated 10 times.
While this provides a duplication-aware upsampling strategy, it is
heavily dataset-dependent -- smaller datasets will have their
distribution of cluster sizes shifted left -- and therefore might not be
scalable across different languages. Additionally, the chosen weights
favor highly duplicated documents the most, which we find are generally of lower quality, and therefore should be repeated less rather than more.

\begin{figure}[h]
    \centering \includegraphics[width=0.9\linewidth]{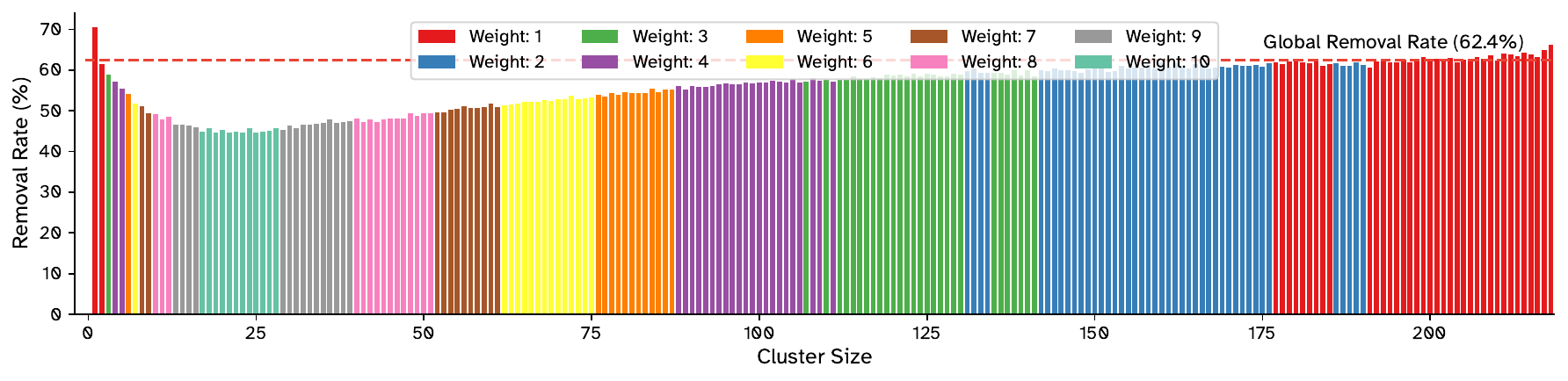}
    \caption{\textbf{Filtering rates by MinHash cluster size} for French documents. The global filtering rate represents the overall percentage of documents removed during the full filtering process. Individual filtering rates are shown for each cluster size, providing a proxy for cluster quality—higher removal rates may indicate lower-quality clusters. We assign \textbf{upsampling weights} to each cluster size based on the filtering rates.}
    \label{fig:rehydration_upsampling_weights}
\end{figure}

While we initially trained models for each of our nine canary languages on
data of different ranges of minhash cluster sizes (e.g., we trained one
model on data that had no duplicates, another on data that had 2
duplicates, data that had 3-4 duplicates, etc) to empirically define
upsampling weights, a simpler and more scalable approach is
to use the results from our filtering stage as a proxy for
cluster size quality: we obtain the global filtering rate
(the percentage of documents removed by our entire filtering process), as
well as the filtering rate for each value of metadata minhash cluster
size, as shown in~\cref{fig:rehydration_upsampling_weights} (for French).

% \begin{figure}[h]
%     \centering \includegraphics[width=0.9\linewidth]{plots/removal_rate_by_cluster_size.pdf}
%     \caption{\textbf{Filtering rates by MinHash cluster size} for French documents. The global filtering rate represents the overall percentage of documents removed during the full filtering process. Individual filtering rates are shown for each cluster size, providing a proxy for cluster quality—higher removal rates may indicate lower-quality clusters.}
%     \label{fig:rehydration_removal_rates}
% \end{figure}

The figure suggests that both data that was never repeated (cluster size of 1), as well as data that is repeated many times
(especially the most-repeated 0.1\% ``long
tail'' of data grouped as the last bar), is generally of lower quality, as our filters removed more
than the global removal rate. Surprisingly, this U looking shape we
observe for French is present in most languages we verified,
but often shifted based on the size of the corpora for each language.

The differences we observe for different cluster sizes align
closely with experimental results from training runs on different
ranges of cluster sizes for the languages we tested, and so we
experimented with setting upsampling weights based on the removal
rates: we assigned a weight of 10 (meaning documents should be repeated
10 times) to the cluster size with the smallest removal rate, and a
weight of 1 to every cluster size above the global removal rate. For the
remaining cluster sizes, we resorted to simple interpolation between
these 2 endpoints. For French, the resulting weights are shown in~\cref{fig:rehydration_upsampling_weights}.
While upsampling weights are dataset-dependent, using the filtering
rates as a proxy for quality is a scalable and affordable method to determine them and
rehydration itself generally provides a strong performance uplift (\cref{fig:all_stages_impact}) with
little downside.

\section{Validating and Applying the FineWeb2 Pipeline}

Having established the pipeline for FineWeb2, and having shown the positive effect of each pipeline step (\cref{fig:all_stages_impact}), we now perform additional evaluations to confirm the effectiveness of our approach and use the pipeline to generate per-language datasets in over 1,000 languages.

% \paragraph{Impact of each stage}\label{sec:impact-each-stage} In \cref{fig:all_stages_impact}, we confirm that the FineWeb2 pipeline provides a performance uplift from each of its stages (language identification filtering, deduplication, heuristic filtering and rehydration) by providing evaluation results of models trained on 350 billion tokens after each stage on Arabic, Chinese, French, Russian, Thai and Turkish. 

\paragraph{Creating the FineWeb2 dataset} We apply our pipeline to 96 Common Crawl snapshots, spanning the summer of 2013 to April 2024, to produce the FineWeb2 dataset, comprising 20 terabytes of text content covering a total of 1,868 language-script pairs, of which 1,226 have over 100 documents, 474 more than 1 thousand documents, and 203 at least 10 thousand documents. Additional details and per-language statistics can be found in~\cref{app:fineweb2-composition}. In addition to the filtered dataset, we also release the preliminary version before filtering is applied, to facilitate further research into alternative filtering methods. As FineWeb2 itself does not include English, for full language coverage we recommend complementing it with FineWeb, whose pipeline inspired FineWeb2.

\paragraph{Comparison to other datasets}
We now compare to other non-English datasets, both on the canary languages used to design the pipeline as well as a set of unseen languages that were not used for ablations.
As discussed previously, prior multilingual datasets often use fixed pipelines across languages, whereas FineWeb2's pipeline adapts to the statistics and characteristics of each language.
By comparing to other multilingual datasets, we can confirm the benefit of FineWeb2's adaptive approach.
To provide a point of comparison against pipelines tuned to a specific language, we additionally evaluate single-language datasets (whose pipelines are designed and tuned for a specific language, often by native speakers) when available.
For canary languages, we use the same set of benchmarks used for pipeline design ablations.
Since the FineWeb2 pipeline was designed specifically around the canary languages, evaluating on unseen languages validates that the pipeline generalizes effectively.
To choose unseen languages, we first followed the same procedure (detailed in \cref{selecting-evaluation-finetasks}) for selecting reliable evaluation tasks across a wide range of languages and chose languages that had a sufficient number of reliable tasks: German, Indonesian, Italian, Japanese and Vietnamese.
The chosen tasks are detailed in \cref{app:unseen}.
Canary-language and unseen-language models were trained for 29 billion and 100 billion tokens respectively.
All evaluated models follow the same architecture, hyperparameters, and (Gemma) tokenizer as considered previously and detailed in \cref{tokenizer-and-model-architecture}.

\begin{figure}[t]
\centering
\includegraphics[width=\textwidth]{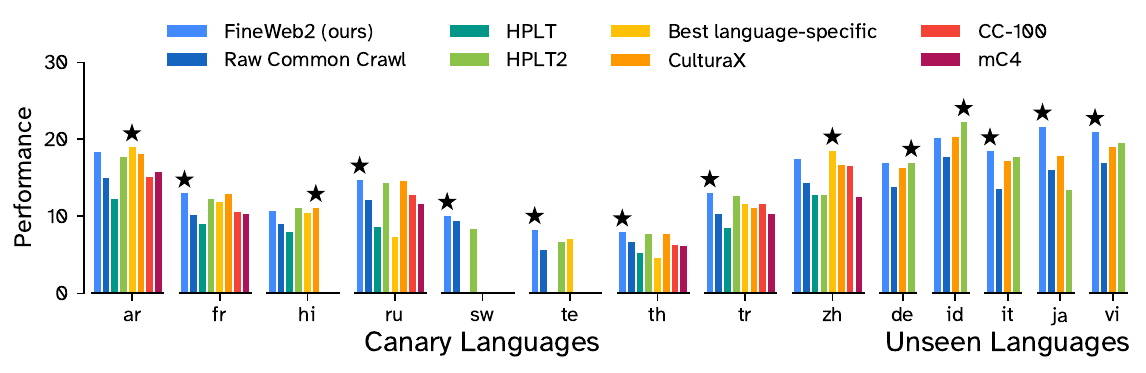}
\vspace{-0.7cm} %
\caption{High-level performance comparison of FineWeb2 to other multilingual and language-specific datasets. We evaluate performance both on the canary languages used to design the FineWeb2 pipeline as well as unseen languages. For brevity, for each language we plot the performance of only the best-performing single-language dataset. The best-performing dataset for each language is marked with $\bigstar$. Expanded results are provided in~\cref{app:canaray-results} and~\cref{app:unseen-results}.}
\label{fig:main_results}
\vspace{-0.45cm} %
\end{figure}

A summary of the results is shown in \cref{fig:main_results}, with detailed per-task results in \cref{app:unseen-results}.
Overall, we found that FineWeb2 produces more performant models than prior multilingal datasets on 11 out of 14 of the languages we considered.
% Interestingly, FineWeb2 underperformed raw Common Crawl on Swahili, possibly due to FineWeb2's filtering producing a limited amount of Swahili data, resulting in overfitting (see \cref{fig:canary_lang_datasets_comparison}).
In some cases, FineWeb2 produces worse performance than a language-specific dataset, which highlights that pipelines hand-designed by language experts can still outperform our adaptive pipeline approach.
These trends hold up both for our canary datasets as well as held-out datasets, which supports the utility of the 1,000+ language-specific datasets we generated with the FineWeb2 pipeline.
On the whole, our results confirm the effectiveness and generalization of our consistent-but-adaptable cross-lingual curation pipeline.

\paragraph{Inspecting low-resource corpora}
% \paragraph{Bible and Wikipedia content in low-resource languages} 
A natural concern is whether low-resource corpora, often with fewer than 20 documents, contain content that is genuinely useful for training.
% A natural concern is whether corpora for languages with only a few documents contain content that is actually useful for training.
% The corpora for many languages consisted of fewer than 20 documents, and so a natural concern is whether these documents contain content that is actually useful for training. 
Manual inspection of over 500 languages reveals that many corpora are composed almost exclusively of Bible and/or Wikipedia content. We categorized the most common document domain names and computed the proportion belonging to Bible- or Wikipedia-related sources: out of 1868 language-script pairs in the final dataset, 70\% (1320 of them) have more than half their documents from Bible- or Wikipedia-related domains. This reflects both the limited availability of online data for many languages and the narrow diversity of sources in the language identifier's training data—where often the only ``clean'' data comes from the Bible~\citep{kargaran-etal-2023-glotlid}. While we hope these corpora remain useful to the research community, their limited diversity highlights the broader challenges of collecting data for the long tail of the world's languages. For more details, see~\cref{app:lower-resource-wiki-bible}.

\section{Conclusion}\label{conclusion}
In this paper, we used a data-driven approach to design a multilingual pre-training data processing pipeline that can automatically adapt to all languages, in contrast to prior work that employs fixed pipelines for each language. We extensively ablate our design choices on a new suite of quantitatively identified multilingual benchmarks that provide a reliable evaluation signal, ultimately covering 14 languages.
We additionally show how duplication counts and filtering results can be leveraged to selectively upsample higher quality content, providing a performance uplift.
Finally, we scaled our pipeline to create FineWeb2, a pre-training dataset covering 1,868 language-script pairs, spanning 20 terabytes of text content curated from 96 Common Crawl snapshots.

While our experiments show that our pipeline yields strong performance, we point out a few limitations. First, although we strove to make the language coverage as wide as possible, computational constraints, language-specific task availability, and excessively small low-resource datasets only enabled us to test a small proportion of the languages in FineWeb2. These factors also forced us to only consider relatively short ablation runs. Second, we studied ``early-signal'' properties of each task at the very early stages of model training, and so it is possible that the properties could change significantly as training progresses, making some tasks more viable. Additionally, we do not explore additional criteria for task selection, such as ``cultural alignment'', with which translated tasks struggle. Similarly, our chosen tasks do not measure other important attributes such as bias or diversity. Lastly, while we strove to include a large number of low-resource languages in our dataset, a large number of them consist almost or even entirely of Bible- or Wikipedia-related content.
Overall, we hope our findings, datasets, and code pave the way for further improvement of datasets that cover a wider range of languages.

\section*{Acknowledgments}
We would like to thank Abdeljalil El Majjodi, Ihssane Nedjaoui, and Zaid Chiech for labeling data for our precision filtering audit; Bram Vanroy, Loïck Bourdois, Omar Kamali, Per Kummervold, Qian Liu, Edwin Rijgersberg, Michael S. Mollel, Faton Rekathati, and Mikhail Tikhomirov for inspecting and providing valuable feedback on their respective native language subsets of FineWeb2; and the many contributors of the FineWeb-C community annotation project.

We extend our gratitude to the Common Crawl project for freely providing and maintaining their regular crawls, which have enabled much of modern LLM research. We thank Pedro Ortiz Suarez from the Common Crawl team, as well as Gema Ramírez, Marta Bañón, and other members of the HPLT team for fruitful discussions about multilingual data.

Additionally, we thank our colleagues -- Nouamane Tazi, Phuc Nguyen, Ferdinand Mom, and Haojun Zhao for designing and building our training framework, Nanotron; Clémentine Fourrier and Nathan Habib for creating and maintaining our evaluation framework, LightEval; and Loubna Ben Allal and Anton Lozhkov for discussions throughout the project. Finally, we thank Hugo Larcher and Mathieu Morlon for tirelessly assisting us whenever we encountered issues with the Hugging Face Science cluster, which they manage with incredible dedication, as well as all the other cluster users for their gracious patience.

% acks:
% - people who labeled data for the wordlist filtering
% - people who gave feedback on data quality for their language
% - HPLT & CommonCrawl people for discussions
% - people who labeled data for the useless fineweb-c (edu thing)
% - colleagues (lighteval team, nanotron team, smollm team, cluster GOATs)

\bibliography{colm2025_conference}
\bibliographystyle{colm2025_conference}

\clearpage
\appendix
\section{Appendix}

\subsection{Word tokenizers for segmentation}
\label{app:word-tokenizers}
For full language coverage, we relied on a wide range of word tokenizers from SpaCy~\citep{Honnibal_spaCy_Industrial-strength_Natural_2020} and Stanza~\citep{qi2020stanzapythonnaturallanguage}, as
well as from libraries targeting specific languages (or language groups)
like IndicNLP~\citep{kunchukuttan2020indicnlp} for Indic Languages, PyThaiNLP~\citep{pythainlp}
for Thai, Kiwipiepy~\citep{kiwipiepy}
for Korean, KhmerNLTK~\citep{hoang-khmer-nltk} for Khmer,
LaoNLP~\citep{wannaphong_phatthiyaphaibun_2022_6833407} for Lao, Botok~\citep{botok} for Tibetan, and
Pyidaungsu~\citep{pyidaungsu} for
Burmese.
For languages without a native word tokenizer, we assigned a proxy
tokenizer from another language based on language family data from the
Ethnologue\footnote{\url{https://www.ethnologue.com/browse/families/}}
using the following approach:

\begin{enumerate}
\item Build a tree for each language family based on the taxonomy from the Ethnologue.
\item Assign tokenizers to each language+script pair that had a native tokenizer from one of the libraries mentioned above
\item Perform an upward pass through the tree, propagating the available tokenizers to the parent nodes, one per script. When multiple tokenizers for the same script are present in the children of a given node, we pick the one from the language with more available data. We do not propagate to the root node as different subfamilies are usually quite different (for example, for Pidgin, ``English-based'' and  ``Swahili-based'' are two branches; for Indo-European, ``Italic'' and ``Armenian'')
\item Perform a downward pass through the tree, assigning as a proxy
  tokenizer the previously propagated parent node tokenizers when
  available.
\end{enumerate}

\begin{figure}[t]
\begin{center}
\begin{tikzpicture}[
  every node/.style={font=\sffamily\small},
  circle node/.style={circle, draw=#1, line width=2.5pt, fill=white, minimum size=1cm, inner sep=0pt},
  square node/.style={rectangle, draw=black!30, fill=#1, minimum size=0.6cm, inner sep=1pt},
  triangle node/.style={regular polygon, regular polygon sides=3, draw=black!30, fill=#1, minimum size=0.8cm, inner sep=0pt},
  edge from parent/.style={draw=black!30, thick}
]
% ROOT NODE - CENTER OF PAGE
\node[circle node=orange] (root) at (5.75,0) {};
\node[anchor=west, xshift=4pt, align=left] at (root.east) {Indo-European\\{[-]} (455)};
% LEVEL 1 - MAXIMALLY SPREAD
\node[circle node=brown!80!white] (germanic) at (0,-3) {};
\node[anchor=west, xshift=4pt, align=left] at (germanic.east) {Germanic\\{[english]} (49)};
\node[circle node=purple!60] (greek) at (3.5,-3) {};
\node[anchor=west, xshift=4pt, align=left] at (greek.east) {Greek\\{[greek]} (6)};
\node[circle node=pink!60] (indoiranian) at (7,-3) {};
\node[anchor=west, xshift=4pt, align=left] at (indoiranian.east) {Indo-Iranian\\{[hindi]} (312)};
\node[circle node=teal!70] (italic) at (10.5,-3) {};
\node[anchor=west, xshift=4pt, align=left] at (italic.east) {Italic\\{[spanish]} (45)};
% LEVEL 2 - WIDER SPACING
\node[circle node=red!80] (latinofaliscan) at (10.5,-6) {};
\node[anchor=west, xshift=4pt, align=left] at (latinofaliscan.east) {Latino-Faliscan\\{[latin]} (1)};
\node[circle node=teal!70] (romance) at (7,-6) {};
\node[anchor=west, xshift=4pt, align=left] at (romance.east) {Romance\\{[spanish]} (44)};
% LEVEL 3 - MAXIMUM HORIZONTAL DISTRIBUTION
\node[triangle node=red!80] (latin) at (10.5,-9) {};
\node[anchor=west, xshift=4pt, align=left] at (latin.east) {Latin\\{[latin]}};
\node[circle node=purple!60] (eastern) at (0,-9) {};
\node[anchor=west, xshift=4pt, align=left] at (eastern.east) {Eastern\\{[romanian]} (4)};
\node[circle node=teal!70] (italowestern) at (7,-9) {};
\node[anchor=west, xshift=4pt, align=left] at (italowestern.east) {Italo-Western\\{[spanish]} (35)};
\node[circle node=teal!70] (southern) at (3.5,-9) {};
\node[anchor=west, xshift=4pt, align=left] at (southern.east) {Southern\\{[spanish]} (5)};
% LEVEL 4 - WELL SPACED
\node[circle node=brown!80!white] (italodalmatian) at (7,-12) {};
\node[anchor=west, xshift=4pt, align=left] at (italodalmatian.east) {Italo-Dalmatian\\{[italian]} (6)};
\node[circle node=teal!70] (western) at (10.5,-12) {};
\node[anchor=west, xshift=4pt, align=left] at (western.east) {Western\\{[spanish]} (29)};
% LEVEL 5 - WIDE SPACING
\node[triangle node=brown!70!orange] (italian) at (3.5,-15) {};
\node[anchor=west, xshift=4pt, align=left] at (italian.east) {Italian\\{[italian]}};
\node[square node=brown!60!orange] (napoletano) at (7,-15) {};
\node[anchor=west, xshift=4pt, align=left] at (napoletano.east) {Napoletano\\{[italian]}};
\node[square node=brown!60!orange] (sicilian) at (10.5,-15) {};
\node[anchor=west, xshift=4pt, align=left] at (sicilian.east) {Sicilian\\{[italian]}};
% Add the ellipsis node for non-displayed families
\node (ellipsis) at (5.75,-1.5) {...};
\draw[edge from parent] (root) -- (ellipsis);
% Connections
\draw[edge from parent] (root) -- (germanic);
\draw[edge from parent] (root) -- (greek);
\draw[edge from parent] (root) -- (indoiranian);
\draw[edge from parent] (root) -- (italic);
\draw[edge from parent] (italic) -- (latinofaliscan);
\draw[edge from parent] (italic) -- (romance);
\draw[edge from parent] (latinofaliscan) -- (latin);
\draw[edge from parent] (romance) -- (eastern);
\draw[edge from parent] (romance) -- (italowestern);
\draw[edge from parent] (romance) -- (southern);
\draw[edge from parent] (italowestern) -- (italodalmatian);
\draw[edge from parent] (italowestern) -- (western);
\draw[edge from parent] (italodalmatian) -- (italian);
\draw[edge from parent] (italodalmatian) -- (napoletano);
\draw[edge from parent] (italodalmatian) -- (sicilian);
% Title at the top
\node[anchor=north, font=\sffamily\bfseries\large] at (5.75,1.5) {Indo-European Language Family Tree};
\end{tikzpicture}
\end{center}
\caption{Example tokenizer assignments based on language family data in Indo-European. Triangles correspond to languages for which a native word tokenizer was available, while squares are languages for which a proxy tokenizer was assigned. The tokenizer assigned to each language is written inside brackets [], and the number of languages in each subnode is in parantheses (). The Italian word tokenizer was propagated to other languages in the Italo-Dalmatian subfamily, while Spanish was propagated up the tree from the Western branch, given that it is a higher resource language than Italian. Latin has its own native tokenizer. Word tokenizers are propagated all the way to the first level, but not to the root or across top level subfamilies.}
\label{fig:tokenizers-family-tree}
\end{figure}
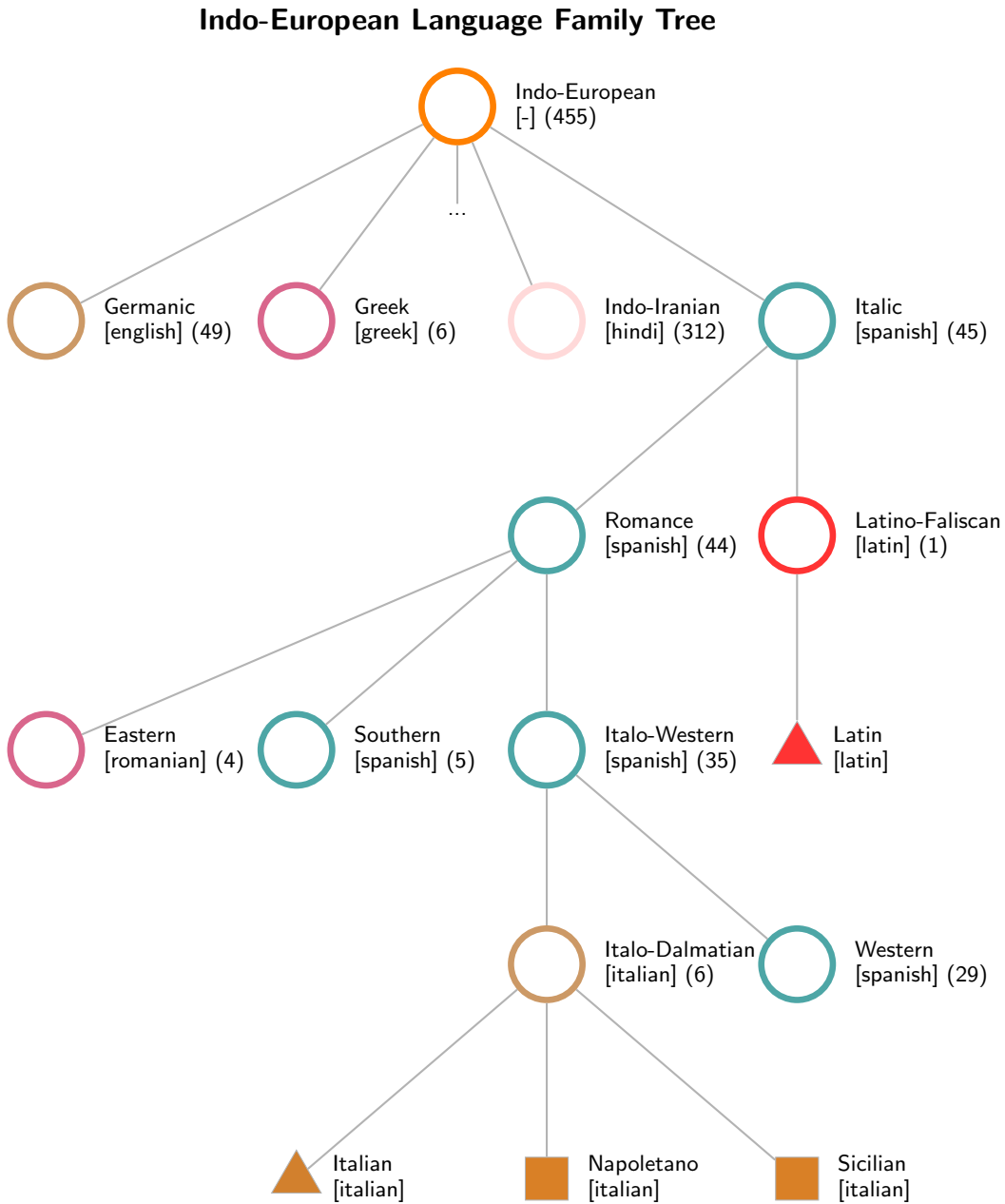

This method allowed us to quickly scale tokenizer assignments for many
languages by assigning tokenizers from a closely related language. An illustrative example is available in~\cref{fig:tokenizers-family-tree}.
We relied on the SpaCy multilingual tokenizer for the remaining languages
with Latin or Cyrillic scripts, which was trained on multiple languages
that use these scripts. For any remaining script, we assigned the remaining languages to the tokenizer of the highest resource language that uses the script and has a native tokenizer.

\clearpage
\subsection{Canary Languages}
While our corpus and pipeline covers more than a thousand languages, we perform in-depth evaluations of the following subset of languages:
\begin{table}[h]
\begin{center}
\begin{tabular}{llll}
\toprule
Language & Family & Script & Resource Availability \\
\midrule
Arabic & Afro-Asiatic & Arabic & Medium \\
Chinese & Sino-Tibetan & Han & High \\
French & Indo-European (Italic) & Latin & High \\
Hindi & Indo-European (Indo-Iranian) & Devanagari & Medium \\
Russian & Indo-European (Balto-Slavic) & Cyrillic & High \\
Swahili & Niger-Congo & Latin & Low \\
Telugu & Dravidian & Telugu & Low \\
Thai & Kra-Dai & Thai & Medium \\
Turkish & Turkic & Latin & Medium \\
\bottomrule
\end{tabular}
\caption{The 9 canary languages and their families, main script, and resource availability.}
\label{tab:canary-languages}
\end{center}
\end{table}

\clearpage
\subsection{Multilingual tokenizers comparison}
\label{app:tokenizer-metrics}

Following~\citet{rust2021goodtokenizermonolingualperformance}, we considered two metrics:

\begin{itemize}
\item
  \textbf{Subword fertility (sf)}: the average number of tokens per ``real'' text
  word. Measures how aggressively a tokenizer splits words. The
  theoretical minimum of 1 would mean the tokenizer vocabulary contains
  every single word from the reference text;
\item
  \textbf{Proportion of continued words (pcw)}: the ratio of ``real'' text
  words encoded with 2 tokens or more. Measures how often a tokenizer
  splits words. A value of 0 means that the tokenizer never splits and 1 that
  it always splits.
\end{itemize}

We split each language's Wikipedia into individual words (see~\cref{sec:separating-words}) and computed the two metrics using tokenizers from a variety of popular multilingual models: Mistral-7B-V3~\citep{jiang2023mistral7b}, Phi3~\citep{abdin2024phi3technicalreporthighly}, Llama3~\citep{llama3modelcard}, Qwen2.5~\citep{qwen2.5}, mT5~\citep{xue2021mt5massivelymultilingualpretrained}, Bigscience-Bloom~\citep{workshop2023bloom176bparameteropenaccessmultilingual}, Command-R~\citep{cohere2024commandr}, Gemma~\citep{gemmateam2024gemmaopenmodelsbased}, and XGLM~\citep{lin2022fewshotlearningmultilinguallanguage}. 
We did not include tokenizers in our comparison if they had a vocabulary size over 256,000, as these would make the embedding layer consume a considerable number of paramaters: at the relatively small model scale we targeted for our experiments (around 1.5 billion parameters), this would force us to significantly reduce the number of model layers due to computational constraints.

Following inspection of the computed metrics in~\cref{tab:tokenizers-comparison}, where we additionally show the average and worst-case (max), and lower is better for both metrics, we excluded tokenizers
that showed very low subword fertility or proportion of
continued words on at least one of our canary languages. The Mistral-7B-V3, Phi3, Command-R and Llama3 tokenizers all do not provide good coverage of Telugu. Additionally, while XGLM and mT5 both showed strong performance, they do not preserve whitespaces, and some characters (particularly for Chinese) would be
encoded as ``unknown token'' ([UNK]). Ultimately, our tokenizer of choice was Gemma, a
modern BPE tokenizer that performed slightly better than
Bigscience-Bloom on average for our experimental setup.

\begin{table}[htbp]
\centering
\label{tab:tokenizers-comparison}
\small
\setlength{\tabcolsep}{3pt}
\begin{tabular}{lccccccccc}
\toprule
Tokenizer & Mistralv3 & Phi3 & Llama3 & Qwen2.5* & mT5 & Bloom & Cmd-R & Gemma & XGLM \\
Vocab size & 32,768 & 100,352 & 128,000 & 151,643 & 250,100 & 250,680 & 255,000 & 256,000 & 256,008 \\
No [UNK] & \ding{51} & \ding{51} & \ding{51} & \ding{51} & \ding{55} & \ding{51} & \ding{51} & \ding{51} & \ding{55} \\
\midrule
English sf & 1.45 & 1.40 & 1.40 & 1.47 & 1.52 & 1.42 & 1.35 & \textbf{1.31} & 1.34 \\
English pcw & 0.23 & 0.28 & 0.28 & 0.29 & 0.45 & 0.31 & 0.22 & \textbf{0.19} & 0.28 \\
\addlinespace
Chinese sf & 3.03 & 2.30 & 1.60 & 1.44 & 2.29 & \textbf{1.29} & 1.35 & 1.43 & 2.21 \\
Chinese pcw & 0.95 & 0.58 & 0.43 & 0.31 & 0.91 & \textbf{0.23} & 0.25 & 0.32 & 0.82 \\
\addlinespace
French sf & 1.69 & 1.74 & 1.73 & 1.76 & 1.71 & 1.49 & 1.50 & 1.50 & \textbf{1.45} \\
French pcw & 0.40 & 0.47 & 0.47 & 0.47 & 0.55 & 0.35 & 0.35 & \textbf{0.34} & 0.35 \\
\addlinespace
Russian sf & 2.42 & 2.99 & 2.34 & 2.50 & 1.96 & 2.86 & 1.99 & 2.05 & \textbf{1.68} \\
Russian pcw & 0.59 & 0.66 & 0.62 & 0.64 & 0.73 & 0.63 & 0.56 & 0.57 & \textbf{0.50} \\
\addlinespace
Turkish sf & 3.18 & 2.63 & 2.32 & 2.55 & 1.99 & 2.59 & 2.13 & 2.22 & \textbf{1.72} \\
Turkish pcw & 0.74 & 0.70 & 0.68 & 0.70 & 0.73 & 0.67 & 0.64 & 0.66 & \textbf{0.53} \\
\addlinespace
Arabic sf & 4.76 & 3.72 & 2.32 & 2.23 & 2.10 & 1.86 & 2.16 & 2.19 & \textbf{1.72} \\
Arabic pcw & 0.92 & 0.86 & 0.74 & 0.67 & 0.79 & 0.60 & 0.68 & 0.69 & \textbf{0.52} \\
\addlinespace
Thai sf & 4.87 & 3.80 & 2.18 & 2.44 & 1.99 & 3.96 & 4.01 & 1.92 & \textbf{1.78} \\
Thai pcw & 0.93 & 0.85 & 0.66 & 0.64 & 0.68 & 0.86 & 0.87 & \textbf{0.46} & 0.53 \\
\addlinespace
Hindi sf & 4.99 & 4.60 & 2.71 & 3.98 & 2.02 & 1.59 & 3.39 & 2.22 & \textbf{1.52} \\
Hindi pcw & 0.91 & 0.90 & 0.81 & 0.86 & 0.69 & 0.39 & 0.80 & 0.60 & \textbf{0.33} \\
\addlinespace
Swahili sf & 2.30 & 2.09 & 2.07 & 2.16 & 1.78 & 1.72 & 1.95 & 1.84 & \textbf{1.54} \\
Swahili pcw & 0.63 & 0.62 & 0.62 & 0.63 & 0.62 & 0.52 & 0.59 & 0.53 & \textbf{0.42} \\
\addlinespace
Telugu sf & 9.83 & 10.11 & 10.11 & 8.41 & 2.44 & \textbf{2.10} & 9.74 & 3.51 & 2.24 \\
Telugu pcw & 0.79 & 0.76 & 0.76 & 0.77 & 0.86 & \textbf{0.59} & 0.78 & 0.74 & 0.69 \\
\midrule
Max sf & 9.83 & 10.11 & 10.11 & 8.41 & 2.44 & 3.96 & 9.74 & 3.51 & \textbf{2.24} \\
Max pcw & 0.95 & 0.90 & 0.81 & 0.86 & 0.91 & 0.86 & 0.87 & \textbf{0.74} & 0.82 \\
\midrule
Avg sf & 4.12 & 3.78 & 3.04 & 3.05 & 2.03 & 2.16 & 3.14 & 2.10 & \textbf{1.76} \\
Avg pcw & 0.76 & 0.71 & 0.64 & 0.63 & 0.73 & 0.54 & 0.61 & 0.55 & \textbf{0.52} \\
\bottomrule
\end{tabular}
\caption{Multilingual Tokenizers Comparison on Wikipedia. * denotes tokenizers that were not originally available when we first ran this comparison. [UNK] is the unknown token: mT5 and XGLM are unable to encode some characters, particularly for Chinese. Avg is the average across all languages, and Max the maximum (worst-case) across all languages. Lower is better for all rows.}
\end{table}

\clearpage
\subsection{Model architecture and training}
\label{app:model-architecture}

\begin{table}[h!]
\centering
\begin{tabular}{ll}
\toprule
Parameter & Value \\
\midrule
Architecture & Llama \\
Number of attention heads & 32 \\
Number of hidden layers & 14 \\
Number of key-value heads & 32 \\
RMS Norm epsilon & 1e-05 \\
d\_model & 2048 \\
Tied word embeddings & True \\
Embedding size & 256008 \\
Total number of parameters & 1.46B \\
Random initialization std & 0.02 \\
Tokenizer & Gemma \\
\bottomrule
\end{tabular}
\label{tab:model-arch}
\caption{\textbf{Architecture configuration} for all models}
\end{table}
% \FloatBarrier

\begin{table}[h!]
\centering
\begin{tabular}{lccc}
\toprule
Parameter & 29BT & 100BT & 350BT \\
\midrule
Data parallelism (dp) & 64 & 56 & 64 \\
Tensor parallelism (tp) & 1 & 1 & 1\\
Pipeline parallelism (pp) & 1 & 1 & 1\\
\midrule
Sequence length & 2048 & 2048 & 2048\\
Batch size (samples) & 1024 & 840 & 1280\\
Batch size (tokens) & 2097152 & 1720320 & 2621440 \\
\bottomrule
\end{tabular}
\label{tab:training-config}
\caption{\textbf{Training settings} for the 3 training scales we consider: 29, 100 and 350 billion tokens. For 100BT and 350BT, we compute critical batch size based on~\citet{deepseekai2024deepseekllmscalingopensource}}
\end{table}
% \FloatBarrier

\begin{table}[h!]
\centering
\begin{tabular}{lccc}
\toprule
Parameter & 29BT & 100BT & 350BT\\
\midrule
Adam beta1 & 0.9 & 0.9 & 0.9\\
Adam beta2 & 0.95 & 0.95 & 0.95\\
Adam epsilon & 1.0e-8 & 1.0e-8 & 1.0e-8 \\
Gradient clipping & 1.0  & 1.0  & 1.0 \\
Weight decay & 0.1 & 0.1 & 0.1\\
\midrule
Learning rate & 3e-4 & 8e-4 & 7e-4\\
Total train steps & 14000 & 59000 & 134000 \\
Warmup steps & 500 & 2950 (5\%) & 6700 (5\%)\\
Warmup style & linear & linear & linear \\
Decay steps & 13500 & 11800 (20\%) & 26800 (20\%)\\
Decay starting step & 500 & 47200 & 107200 \\
Decay style & cosine & linear & linear\\
Minimum decay LR & 3.0e-5 & 0 & 0\\
\bottomrule
\end{tabular}
\label{tab:optimizer-config}
\caption{\textbf{Optimizer settings} for the 3 training scales we consider: 29, 100 and 350 billion tokens. For 100BT and 350BT, we train with a constant learning rate until the last 20\% of steps (computed following~\citet{deepseekai2024deepseekllmscalingopensource}), so that the resulting models can easily undergo continued pretraining.}
\end{table}
% \FloatBarrier

\clearpage
\subsection{Evaluation details}
\label{app:Fine-Tasks}

\subsubsection{Task selection criteria}
\label{app:Fine-Tasks criteria}
As noted in~\cref{selecting-evaluation-finetasks}, we define precise quantitative criteria for each of the properties of the early-signal task. To compute each criterion, we only use models trained on available reference datasets for given language (see~\cref{sec:reference-datasets}), denoted as $M$. Every task in our final selection had to satisfy all of the following criteria requirements:

\paragraph{Monotonicity.}  
To assess \textit{Monotonicity} of a task, we compute the average Spearman rank correlation between the evaluation steps and the corresponding model scores. For a given model \( m \), let the score at step \( s \) be denoted \( m(s) \). The average monotonicity across all models is then defined as:

\[
  \bar{\rho} = \frac{1}{|M|} \sum_{m \in M} \rho\left([s_0, s_1, \ldots, s_n], [m(s_0), m(s_1), \ldots, m(s_n)]\right)
\]

Here, the Spearman correlation \( \rho(x, y) \) between sequences \( x = [x_1, \ldots, x_n] \) and \( y = [y_1, \ldots, y_n] \) is computed as:

\[
  \rho(x, y) = 1 - \frac{6 \sum_{i=1}^{n} d_i^2}{n(n^2 - 1)}
\]

where \( d_i = \text{rank}(x_i) - \text{rank}(y_i) \) is the rank difference for element \( i \), and \( n \) is the number of evaluation steps.  
We consider a task to meet the monotonicity criterion if:

\[
\bar{\rho} \geq 0.5
\]

\paragraph{Signal-to-Noise Ratio (SNR).}  
Inspired by~\cite{variance_meta} we estimate how robust is a task to training noise, by computing its \textit{Signal-to-Noise Ratio (SNR)} using four models trained on unfiltered CommonCrawl data under different random seeds:

\begin{itemize}
\item \textbf{seed-3:} Trained on a random subset with data and model seed set to 3
\item \textbf{seed-4:} Trained on same subset as seed 3, with data and model seed set to 4
\item \textbf{seed-5:} Trained on a different random subset with data and model seed set to 5
\item \textbf{seed-6:} Trained on the same subset as 5 with data seed = 6 and model seed = 42
 
\end{itemize}

We refer to this set of four models as $MC$. For each evaluation step \( s \), we define the mean score (signal) as:

\[
\mu^*_s = \frac{1}{|MC|} \sum_{m \in MC} m(s)
\]

and the standard deviation (noise) as:

\[
\sigma_s = \sqrt{ \frac{1}{|MC|} \sum_{m \in MC} \left( m(s) - \mu^*_s \right)^2 }
\]

The overall task SNR is then the average ratio of signal to noise across all $n$ training steps:

\[
\mathrm{SNR} = \frac{1}{n} \sum_{s=0}^{n} \frac{\mu^*_{s}}{\sigma_s}
\]

We chose the minimum required SNR to 20, with the exception of generative tasks, which we found to be considerably "noisier" in general, but we wanted to have at least one 
generative task per language. Generative tasks are quite relevant in a multilingual
context as they provide insights into how the model behaves when
prompted to generate unconstrained, i.e., without a limited set of
answer options. Models trained in multiple languages can sometimes
exhibit high scores in multiple choice tasks but reply in the wrong
language for generative tasks (``accidental translation''), or otherwise
lack fluency~\citep{xue2021mt5massivelymultilingualpretrained}. 

\paragraph{Non-Random Performance.}  
To assess that non-zero task results are not just a consequence of random noise, we look at the best score at the last evaluation step among models from $M$.
We first compute the maximum improvement over a random baseline \( b \):

\[
\text{max}_d = \max_{m \in M} \left( m(n) - b \right)
\]

We then estimate the variance at the end of training using the standard deviation (from previous calculation) averaged over the last 5 steps:

\[
\sigma_{\text{end}} = \frac{1}{5} \sum_{s=n-4}^{n} \sigma_s
\]

Finally, The non-randomness score is defined as the ratio of max improvement to this terminal variance:

\[
\mathrm{non\_randomness} = \frac{\text{max}_d}{\sigma_{\text{end}}}
\]

A task satisfies the non-randomness criterion if:

\[
\mathrm{non\_randomness} \geq 3
\]

\paragraph{Ordering Consistency}
To compute how consistently models are ordered as training progresses, we calculate the average Kendall Tau-a between model rankings at consecutive steps in the second half of training. We ignore the first 15 billion tokens, as we are interested in this property at a later stage of training, and in the first half, we found the ordering to be very inconsistent, skewing the overall score. First, we define Kendall Tau-a of model ranking as:
\[
\tau_a(x, y) = \frac{C - D}{\binom{n}{2}}
\]

where \(C\) and \(D\) are the number of concordant and discordant pairs between the rankings \(x\) and \(y\) of the model scores at steps \(s_i\) and \(s_{i+1}\). The overall consistency is:

\[
\mathrm{ordering\_consistency} = \frac{1}{|P|} \sum_{(s_i, s_{i+1}) \in P} \tau_a\left(r(s_i), r(s_{i+1})\right)
\]

where \(P\) is the set of consecutive step pairs in the latter half of the training, and \(r(s)\) is the ranking of model scores in step \(s\).

While we first considered using the criteria for selection, we could not determine a reliable threshold for the criterion and therefore only use it for observational reasons.

\subsubsection{Metrics and Formulation}
% the following paragraphs can potentially go in the appendix
For non-generative tasks, we compute accuracies using \textbf{Cloze Formulation} (CF, completing with the full option
text) in place of the more commonly used
Multi-Choice Formulation (MCF, completing with A/B/C/D), as previous
work has shown that MCF has random performance in the early stages of
training~\citep{gu2025olmesstandardlanguagemodel,li2024datacomplmsearchgenerationtraining}.

Additionally, since all models that we compare use the same tokenizer,
we \textbf{normalize answer log-probabilities based on token count} instead of
number of characters, and use pointwise mutual information
(\textbf{PMI})~\citep{gu2025olmesstandardlanguagemodel} for more difficult tasks such as AGIEval~\citep{zhong2023agievalhumancentricbenchmarkevaluating} or translated versions of MMLU~\citep{hendryckstest2021}.
For these tasks, we use the
\textbf{F1-score} of overlapping words, as it is generally less noisy
and more resilient to small changes in the generations than exact
matching (which in turn might be more appropriate for math related
tasks, which we do not evaluate on).

\clearpage
\subsubsection{List of selected evaluation tasks for canary languages}
\label{app:finetasks-selection}

\begin{table}[h]
\begin{center}
\small
\setlength{\tabcolsep}{3pt}
\begin{tabular}{lcccccc}
\toprule
Task & Type & Metric & Mono & SNR & Rand & Order \\
\midrule
\text{Belebele} \citep{alghafa} & RC & Acc (Char) & 0.61 & 58.23 & 14.67 & 0.13 \\
\text{ArabicMMLU} \citep{arabic_mmlu} & GK & Acc (PMI) & 0.81 & 80.00 & 18.28 & 0.91 \\
\text{X-CSQA} \citep{xcsr} & RES & Acc (PMI) & 0.65 & 33.44 & 11.13 & 0.91 \\
\text{Alghafa: MCQ Exams} \citep{alghafa} & GK & Acc (Token) & 0.51 & 35.49 & 8.89 & 0.61 \\
\text{Alghafa: SOQAL} \citep{alghafa} & RC & Acc (Token) & 0.74 & 46.22 & 33.78 & 0.11 \\
\text{Alghafa: ARC Easy} \citep{oall2024} & GK & Acc (Token) & 0.74 & 76.58 & 35.41 & 0.91 \\
\text{Okapi: Hellaswag} \citep{okapi} & NLU & Acc (Token) & 0.80 & 43.05 & 12.01 & 0.97 \\
\text{OALL2024: PIQA} \citep{oall2024} & RES & Acc (Token) & 0.81 & 69.34 & 7.69 & 0.71 \\
\text{OALL2024: RACE} \citep{oall2024} & RC & Acc (Token) & 0.82 & 66.01 & 18.22 & 0.43 \\
\text{OALL2024: SCIQ} \citep{oall2024} & GK & Acc (Token) & 0.80 & 74.06 & 32.87 & 0.70 \\
\text{X-CODAH} \citep{xcsr} & RES & Acc (Token) & 0.75 & 24.80 & 8.50 & 0.31 \\
\text{X-Story Cloze} \citep{xstory_cloze} & NLU & Acc (Token) & 0.87 & 93.20 & 9.76 & 0.83 \\
\text{ARCD} \citep{arcd} & GK & F1 & 0.83 & 28.28 & 35.58 & 0.83 \\
\text{MLQA} \citep{mlqa} & RC & F1 & 0.86 & 17.27 & 24.83 & 0.87 \\
\text{Tydiqa} \citep{tydiqa} & RC & F1 & 0.86 & 27.17 & 55.07 & 0.94 \\
\bottomrule
\end{tabular}
\caption{Selected tasks for Arabic satisfying the early-signal conditions: Monotonicity (Mono), Signal-to-noise ratio (SNR), Non-Randomness (Rand) and Ordering Consistency (Order)}
\label{tab:task_selection_ar}
\end{center}
\end{table}

%zh finetasks
\begin{table}[h]
\begin{center}
\small
\setlength{\tabcolsep}{3pt}
\begin{tabular}{lcccccc}
\toprule
Task & Type & Metric & Mono & SNR & Rand & Order \\
\midrule
\text{AGIEval (ZH subset)} \citep{agieval} & GK & Acc (PMI) & 0.46 & 98.82 & 15.86 & 0.86 \\
\text{X-CSQA} \citep{xcsr} & RES & Acc (PMI) & 0.83 & 25.63 & 10.09 & 0.89 \\
\text{Belebele} \citep{belebele} & RC & Acc (Token) & 0.51 & 74.30 & 15.36 & 0.70 \\
\text{C3} \citep{c3} & RC & Acc (Token) & 0.87 & 72.89 & 36.01 & 0.66 \\
\text{C-Eval} \citep{ceval} & GK & Acc (Token) & 0.75 & 50.20 & 8.04 & 0.53 \\
\text{CMMLU} \citep{cmmlu} & GK & Acc (Token) & 0.91 & 117.92 & 21.93 & 0.96 \\
\text{Okapi: Hellaswag} \citep{okapi} & NLU & Acc (Token) & 0.87 & 70.60 & 21.95 & 0.97 \\
\text{M3Exam} \citep{m3exam} & GK & Acc (Token) & 0.74 & 36.02 & 8.75 & 0.67 \\
\text{X-CODAH} \citep{xcsr} & RES & Acc (Token) & 0.66 & 32.65 & 14.72 & 0.66 \\
\text{X-COPA} \citep{xcopa} & RES & Acc (Token) & 0.80 & 77.20 & 15.06 & 0.69 \\
\text{X-Story Cloze} \citep{xstory_cloze} & NLU & Acc (Token) & 0.87 & 79.20 & 15.57 & 0.84 \\
\text{X-Winograd} \citep{xwinograd} & NLU & Acc (Token) & 0.88 & 102.87 & 21.83 & 0.86 \\
\text{Chinese SQuAD} \citep{chinese_squad} & RC & F1 & 0.85 & 27.71 & 27.40 & 0.90 \\
\text{CMRC} \citep{cmrc} & RC & F1 & 0.91 & 25.33 & 34.43 & 0.67 \\
\text{MLQA} \citep{mlqa} & RC & F1 & 0.91 & 23.76 & 20.40 & 0.86 \\
\bottomrule
\end{tabular}
\caption{Selected tasks for Chinese satisfying the early-signal conditions: Monotonicity (Mono), Signal-to-noise ratio (SNR), Non-Randomness (Rand) and Ordering Consistency (Order)}
\label{tab:task_selection_zh}
\end{center}
\end{table}

%French Fine Tasks
\begin{table}[h]
\begin{center}
\small
\setlength{\tabcolsep}{3pt}
\begin{tabular}{lcccccc}
\toprule
Task & Type & Metric & Mono & SNR & Rand & Order \\
\midrule
\text{Okapi: ARC} \citep{okapi} & GK & Acc (PMI) & 0.69 & 30.10 & 3.33 & 0.47 \\
\text{Meta MMLU} \citep{grattafiori2024llama3herdmodels} & GK & Acc (PMI) & 0.87 & 107.58 & 10.95 & 0.56 \\
\text{X-CSQA} \citep{xcsr} & RES & Acc (PMI) & 0.83 & 30.50 & 11.01 & 0.76 \\
\text{Belebele} \citep{belebele} & RC & Acc (Token) & 0.85 & 33.68 & 5.65 & 0.39 \\
\text{Okapi: Hellaswag} \citep{okapi} & NLU & Acc (Token) & 0.96 & 71.11 & 30.84 & 0.70 \\
\text{X-CODAH} \citep{xcsr} & RES & Acc (Token) & 0.74 & 33.68 & 9.19 & 0.74 \\
\text{FQuad} \citep{fquad} & RC & F1 & 0.91 & 14.64 & 19.08 & 0.69 \\
\text{Mintaka} \citep{mintaka} & GK & F1 & 0.82 & 6.91 & 12.92 & 0.79 \\
\bottomrule
\end{tabular}
\caption{Selected tasks for French satisfying the early-signal conditions: Monotonicity (Mono), Signal-to-noise ratio (SNR), Non-Randomness (Rand) and Ordering Consistency (Order)}
\label{tab:task_selection_fr}
\end{center}
\end{table}

%Hindi Finetask
\begin{table}[h]
\begin{center}
\small
\setlength{\tabcolsep}{3pt}
\begin{tabular}{lcccccc}
\toprule
Task & Type & Metric & Mono & SNR & Rand & Order \\
\midrule
\text{Meta MMLU} \citep{grattafiori2024llama3herdmodels} & GK & Acc (PMI) & 0.68 & 97.78 & 9.13 & 0.33 \\
\text{X-CSQA} \citep{xcsr} & RES & Acc (PMI) & 0.60 & 22.84 & 4.45 & 1.00 \\
\text{Belebele} \citep{belebele} & RC & Acc (Token) & 0.61 & 66.05 & 6.65 & 0.76 \\
\text{Okapi: Hellaswag} \citep{okapi} & NLU & Acc (Token) & 0.87 & 47.47 & 16.35 & 1.00 \\
\text{Okapi: ARC} \citep{okapi} & GK & Acc (Token) & 0.95 & 62.19 & 23.11 & 0.67 \\
\text{X-CODAH} \citep{xcsr} & RES & Acc (Token) & 0.53 & 39.83 & 14.13 & 0.67 \\
\text{X-Story Cloze} \citep{xstory_cloze} & NLU & Acc (Token) & 0.74 & 87.75 & 8.39 & 1.00 \\
\text{IndicQA} \citep{indicqa} & RC & F1 & 0.94 & 13.20 & 12.20 & 0.81 \\
\bottomrule
\end{tabular}
\caption{Selected tasks for Hindi satisfying the early-signal conditions: Monotonicity (Mono), Signal-to-noise ratio (SNR), Non-Randomness (Rand) and Ordering Consistency (Order)}
\label{tab:task_selection_hi}
\end{center}
\end{table}

%Russian FineTasks
\begin{table}[h]
\begin{center}
\small
\setlength{\tabcolsep}{3pt}
\begin{tabular}{lcccccc}
\toprule
Task & Type & Metric & Mono & SNR & Rand & Order \\
\midrule
\text{Okapi: ARC} \citep{okapi} & GK & Acc (PMI) & 0.55 & 35.17 & 3.76 & 0.53 \\
\text{RUMMLU} \citep{mera} & GK & Acc (PMI) & 0.77 & 64.24 & 6.10 & 0.56 \\
\text{X-CSQA} \citep{xcsr} & RES & Acc (PMI) & 0.73 & 38.45 & 16.03 & 0.71 \\
\text{Belebele} \citep{belebele} & RC & Acc (Token) & 0.81 & 61.97 & 19.26 & 0.71 \\
\text{Okapi: Hellaswag} \citep{okapi} & NLU & Acc (Token) & 0.97 & 86.76 & 28.22 & 0.83 \\
\text{Parus} \citep{mera} & RES & Acc (Token) & 0.93 & 81.06 & 24.61 & 0.67 \\
\text{OpenBookQA} \citep{mera} & RES & Acc (Token) & 0.73 & 43.43 & 18.08 & 0.73 \\
\text{X-CODAH} \citep{xcsr} & RES & Acc (Token) & 0.85 & 26.97 & 6.79 & 0.50 \\
\text{X-Story Cloze} \citep{xstory_cloze} & NLU & Acc (Token) & 0.93 & 66.81 & 12.04 & 0.84 \\
\text{Sber SQuAD} \citep{sber_squad} & RC & F1 & 0.89 & 9.93 & 10.85 & 0.84 \\
\text{Tydiqa} \citep{tydiqa} & RC & F1 & 0.92 & 10.44 & 11.28 & 0.83 \\
\text{X-QuAD} \citep{xquad} & RC & F1 & 0.90 & 8.79 & 7.56 & 0.60 \\
\bottomrule
\end{tabular}
\caption{Selected tasks for Russian satisfying the early-signal conditions: Monotonicity (Mono), Signal-to-noise ratio (SNR), Non-Randomness (Rand) and Ordering Consistency (Order)}
\label{tab:task_selection_ru}
\end{center}
\end{table}

\begin{table}[h]
\begin{center}
\small
\setlength{\tabcolsep}{3pt}
\begin{tabular}{lcccccc}
\toprule
Task & Type & Metric & Mono & SNR & Rand & Order \\
\midrule
\text{Okapi: ARC} \citep{okapi} & GK & Acc (Token) & 0.88 & 60.69 & 6.32 & - \\
\text{Belebele} \citep{belebele} & RC & Acc (Token) & 0.44 & 65.26 & 5.44 & - \\
\text{M3Exam} \citep{m3exam} & GK & Acc (Token) & 0.63 & 34.82 & 3.52 & - \\
\text{X-COPA} \citep{xcopa} & RES & Acc (Token) & 0.82 & 74.71 & 4.66 & - \\
\text{X-Story Cloze} \citep{xstory_cloze} & NLU & Acc (Token) & 0.86 & 130.08 & 20.54 & - \\
\text{KenSWQuAD} \citep{kenswquad} & RC & F1 & 0.91 & 12.95 & 12.43 & - \\
\text{Tydiqa} \citep{tydiqa} & RC & F1 & 0.65 & 12.67 & 15.01 & - \\
\bottomrule
\end{tabular}
\caption{Selected tasks for Swahili satisfying the early-signal conditions: Monotonicity (Mono), Signal-to-noise ratio (SNR), Non-Randomness (Rand) and Ordering Consistency (Order)}
\label{tab:task_selection_sw}
\end{center}
\end{table}

%Telgu finetasks
\begin{table}[h]
\begin{center}
\small
\setlength{\tabcolsep}{3pt}
\begin{tabular}{lcccccc}
\toprule
Task & Type & Metric & Mono & SNR & Rand & Order \\
\midrule
\text{Okapi: Hellaswag} \citep{okapi} & NLU & Acc (Token) & 0.82 & 56.06 & 7.84 & - \\
\text{Okapi: MMLU} \citep{okapi} & GK & Acc (Token) & 0.92 & 148.57 & 4.11 & - \\
\text{X-COPA} \citep{xcopa} & RES & Acc (Token) & 0.77 & 69.31 & 6.01 & - \\
\text{X-Story Cloze} \citep{xstory_cloze} & NLU & Acc (Token) & 0.67 & 108.25 & 8.02 & - \\
\text{IndicQA} \citep{indicqa} & RC & F1 & 0.72 & 12.39 & 9.65 & - \\
\bottomrule
\end{tabular}
\caption{Selected tasks for Telugu satisfying the early-signal conditions: Monotonicity (Mono), Signal-to-noise ratio (SNR), Non-Randomness (Rand) and Ordering Consistency (Order)}
\label{tab:task_selection_te}
\end{center}
\end{table}

%Thai Finetasks
\begin{table}[h]
\begin{center}
\small
\setlength{\tabcolsep}{3pt}
\begin{tabular}{lcccccc}
\toprule
Task & Type & Metric & Mono & SNR & Rand & Order \\
\midrule
\text{Meta MMLU} \citep{grattafiori2024llama3herdmodels} & GK & Acc (PMI) & 0.54 & 93.51 & 6.42 & 0.60 \\
\text{Belebele} \citep{belebele} & RC & Acc (Token) & 0.63 & 53.88 & 13.65 & 0.66 \\
\text{Translated Hellaswag} \citep{hellaswag_th} & NLU & Acc (Token) & 0.69 & 52.78 & 11.51 & 0.53 \\
\text{M3Exam} \citep{m3exam} & GK & Acc (Token) & 0.75 & 45.32 & 4.24 & 0.50 \\
\text{ThaiQA} \citep{thaiqa} & RC & F1 & 0.90 & 20.39 & 15.92 & 0.66 \\
\text{X-QuAD} \citep{xquad} & RC & F1 & 0.90 & 17.45 & 20.07 & 0.80 \\
\bottomrule
\end{tabular}
\caption{Selected tasks for Thai satisfying the early-signal conditions: Monotonicity (Mono), Signal-to-noise ratio (SNR), Non-Randomness (Rand) and Ordering Consistency (Order)}
\label{tab:task_selection_th}
\end{center}
\end{table}

% tr finetasks
\begin{table}[h]
\begin{center}
\small
\setlength{\tabcolsep}{3pt}
\begin{tabular}{lcccccc}
\toprule
Task & Type & Metric & Mono & SNR & Rand & Order \\
\midrule
\text{TR Leaderboard: ARC} \citep{tr_leaderboard} & GK & Acc (Char) & 0.91 & 49.33 & 21.32 & 0.79 \\
\text{Belebele} \citep{belebele} & RC & Acc (Char) & 0.50 & 47.97 & 5.93 & 0.09 \\
\text{Exams} \citep{exams} & GK & Acc (Char) & 0.78 & 31.73 & 5.96 & 0.33 \\
\text{Okapi: Hellaswag} \citep{okapi} & NLU & Acc (Char) & 0.95 & 58.56 & 21.45 & 0.90 \\
\text{X-COPA} \citep{xcopa} & RES & Acc (Char) & 0.61 & 81.18 & 11.43 & 0.66 \\
\text{TR Leaderboard: MMLU} \citep{tr_leaderboard} & GK & Acc (PMI) & 0.81 & 95.48 & 12.60 & 0.61 \\
\text{THQuAD} \citep{thquad} & RC & F1 & 0.93 & 17.06 & 20.03 & 0.60 \\
\text{X-QuAD} \citep{xquad} & RC & F1 & 0.92 & 26.33 & 28.74 & 0.73 \\
\bottomrule
\end{tabular}
\caption{Selected tasks for Turkish satisfying the early-signal conditions: Monotonicity (Mono), Signal-to-noise ratio (SNR), Non-Randomness (Rand) and Ordering Consistency (Order)}
\label{tab:task_selection_tr}
\end{center}
\end{table}

\clearpage
\subsection{Language Identification}
\subsubsection{Classifier choice}
\label{app:classifier-choice}

While Transformer-based LID classifiers exist~\citep{bapna2022buildingmachinetranslationsystems}, they are too slow
and expensive to run at a large scale. Most commonly used LID classifiers
are simple models based on character level n-grams like CLD3~\citep{salcianu2018compact} (107 supported languages),
used in mC4, or classifiers following the fastText architecture~\citep{joulin2016fasttextzipcompressingtextclassification}, such as FT176
(176 languages) used in CC-100~\citep{wenzek-etal-2020-ccnet,conneau-etal-2020-unsupervised}, and CulturaX~\citep{nguyen-etal-2024-culturax}, as well as in many
English-only datasets~\citep{soldaini2024dolmaopencorpustrillion, penedo2023refinedwebdatasetfalconllm}; OpenLID~\citep{burchell-etal-2023-open} (193
languages), used in HPLT2~\citep{burchell2025expandedmassivemultilingualdataset}; and the recent GlotLID~\citep{kargaran-etal-2023-glotlid} (1880 languages).

We used the GlotLID~\citep{kargaran-etal-2023-glotlid}, specifically version V3—the latest available at the time of our experiments~\citep{kargaranglotcc}. This LID classifier covers a large number of languages and addresses some common issues in LID classifiers:

\begin{itemize}
\item
  Its large language coverage can reduce ``out-of-model cousin'' errors~\citep{caswell-etal-2020-language,bapna2022buildingmachinetranslationsystems}, where unsupported
  languages can be misclassified as a closely related supported
  language
\item
  It explicitly distinguishes scripts (Latin, Arabic, Cyrillic, etc), improving detection for languages that support multiple scripts
\item
  It provides different labels based on script e.g.~`arb\_Arab` is
  Standard Arabic in Arabic script, while `arb\_Latn` is Standard Arabic
  in Latin script, allowing us to tailor the filtering to each script
\item
  It includes an ``UND'' label for non supported scripts, so that
  languages that use them aren't misclassified as supported languages
\item
  Includes specific labels trained on ``noise'' documents, such as text
  decoded with the wrong encoding, binary content, or misrendered PDFs,
  preventing it from being classified as a natural text language
\end{itemize}

\begin{figure}[h]
    \centering \includegraphics[width=0.9\linewidth]{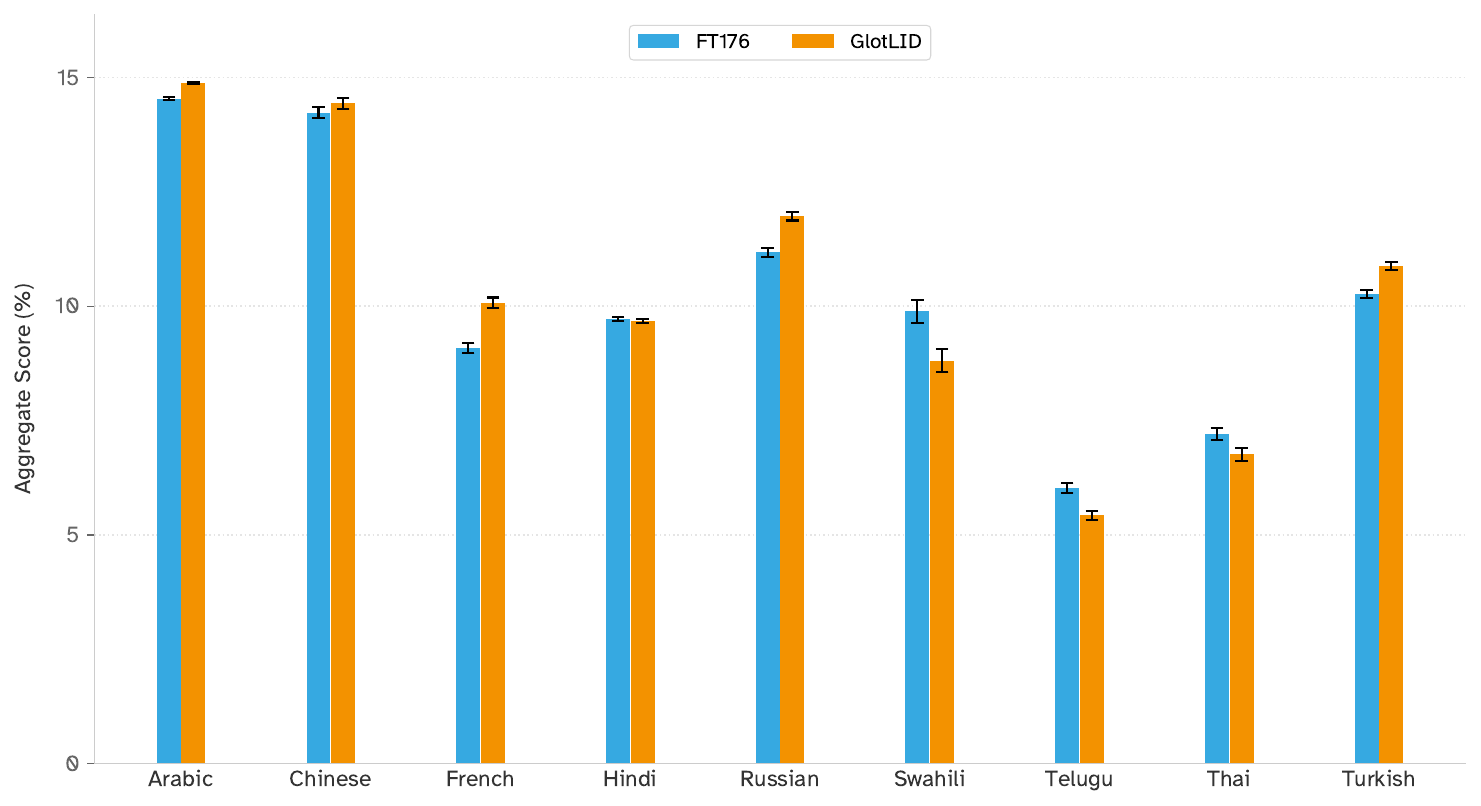}
    \caption{\textbf{FT176 vs GlotLID} without any threshold filtering applied to either classifier. While GlotLID seems to outperform in higher resource languages, FT176 performs slightly better on lower resource languages. However, GlotLID supports a considerably larger number of (lower-resource) languages.}
    \label{fig:ft176vsglotlid}
\end{figure}

In~\cref{fig:ft176vsglotlid} we present a comparison between GlotLID and FT176, without any threshold filtering.

\clearpage
\subsubsection{Confidence Threshold}

For each language, we set thresholds at regular removal intervals (threshold to remove 5\% of data, 10\% of data, etc) and at other values of interest (e.g., 0.7, 0.9). We then train models on 30 billion tokens of the filtered data using each threshold and evaluate the resulting models. In~\cref{tab:lang_threshold_ar,tab:lang_threshold_zh,tab:lang_threshold_fr,tab:lang_threshold_hi,tab:lang_threshold_ru,tab:lang_threshold_sw,tab:lang_threshold_te,tab:lang_threshold_th,tab:lang_threshold_tr}, the highest scoring range of thresholds are marked in bold. Our formula to automatically set thresholds based on the mean and standard deviation of each language's confidence scores distribution selects values within the highest scoring range for all languages except Chinese and Hindi (\cref{tab:language_threshold_parameters}).

\begin{table}[h]
\centering
\begin{tabular}{lcccc}
\toprule
Language & Minimum & Maximum & Formula & Value in range \\
\midrule
Arabic & 0.883 & 0.9 & 0.8812 & \ding{51}  \\
Chinese & 0.895 & 0.937 & 0.7415 & \ding{55} \\
French & 0.750 & 0.932 & 0.8195 & \ding{51} \\
% need to double check this one as it's a mess
Hindi & 0.483 & 0.557 & 0.6827 & \ding{55}\\
Russian & 0.8 & - & 0.9 & \ding{51} \\
Swahili & 0.186 & 0.544 & 0.3 & \ding{51} \\
Telugu & 0.701 & 0.701 & 0.7002 & \ding{51} \\
Thai & 0.9 & 0.961 & 0.9 & \ding{51}  \\
Turkish & 0.866 & - & 0.8753 & \ding{51} \\
\bottomrule
\end{tabular}
\caption{Minimum and Maximum refer to the highest performing threshold range endpoints. Formula is the value defined by our threshold setting formula, which we ultimately use for all languages.}
\label{tab:language_threshold_parameters}
\end{table}

% ar
\begin{table}[h]
\begin{center}
\small
\setlength{\tabcolsep}{3pt}
\begin{tabular}{lccccc}
\toprule
 & Model 1 & Model 2 & Model 3 & Model 4 & Model 5 \\\midrule
Threshold & 0.000 & 0.700 & 0.883 & 0.900 & 0.968 \\
\% Removed & 0.0\% & 3.0\% & 5.0\% & 5.4\% & 10.0\% \\
Aggregate Score & 14.9\% & 15.2\% & \textbf{15.4\%} & \textbf{16.1\%} & 15.2\%\\
\bottomrule
\end{tabular}
\end{center}
\caption{Arabic Threshold Analysis}
\label{tab:lang_threshold_ar}
\end{table}

% zh
\begin{table}[h]
\begin{center}
\small
\setlength{\tabcolsep}{3pt}
\begin{tabular}{lccccccc}
\toprule
 & Model 1 & Model 2 & Model 3 & Model 4 & Model 5 & Model 6 & Model 7 \\\midrule
Threshold & 0.000 & 0.214 & 0.429 & 0.678 & 0.806 & 0.895 & 0.937 \\
\% Removed & 0.0\% & 5.0\% & 10.0\% & 15.0\% & 20.0\% & 25.0\% & 30.0\% \\
Aggregate Score & 14.4\% & 14.4\% & 14.8\% & 14.8\% & 14.4\% & \textbf{15.1\%} & \textbf{15.1\%} \\
\bottomrule
\end{tabular}
\end{center}
\caption{Chinese Threshold Analysis}
\label{tab:lang_threshold_zh}
\end{table}

% fr
\begin{table}[h]
\begin{center}
\small
\setlength{\tabcolsep}{3pt}
\begin{tabular}{lccccccc}
\toprule
 & Model 1 & Model 2 & Model 3 & Model 4 & Model 5 & Model 6 & Model 7 \\\midrule
Threshold & 0.000 & 0.467 & 0.723 & 0.750 & 0.800 & 0.867 & 0.932 \\
\% Removed & 0.0\% & 5.0\% & 10.0\% & 10.7\% & 12.2\% & 15.0\% & 20.0\% \\
Aggregate Score & 10.1\% & 10.5\% & 10.5\% & \textbf{11.1\% }& \textbf{10.7\%} & \textbf{11.0\% }& \textbf{11.2\%} \\
\bottomrule
\end{tabular}
\end{center}
\caption{French Threshold Analysis}
\label{tab:lang_threshold_fr}
\end{table}

% hi
\begin{table}[h]
\begin{center}
\small
\setlength{\tabcolsep}{4pt}
\begin{tabular}{lccccc}
\toprule
 & Model 1 & Model 2 & Model 3 & Model 4 & Model 5 \\\midrule
Threshold & 0.000 & 0.483 & 0.557 & 0.616 & 0.669 \\
\% Removed & 0.0\% & 5.0\% & 10.0\% & 15.0\% & 20.0\% \\
Aggregate Score & 9.7\% & \textbf{9.8\%} & \textbf{9.8\%} & 9.2\% & 9.3\% \\
\bottomrule
\end{tabular}

\vspace{0.2cm}

\begin{tabular}{lcccccc}
\toprule
 & Model 6 & Model 7 & Model 8 & Model 9 & Model 10 & Model 11 \\\midrule
Threshold & 0.714 & 0.752 & 0.786 & 0.815 & 0.840 & 0.862 \\
\% Removed & 25.0\% & 30.0\% & 35.0\% & 40.0\% & 45.0\% & 50.0\% \\
Aggregate Score & 9.5\% & 9.7\% & 9.0\% & 9.3\% & 9.2\% & 8.7\%\\
\bottomrule
\end{tabular}
\end{center}
\caption{Hindi Threshold Analysis}
\label{tab:lang_threshold_hi}
\end{table}

% ru
\begin{table}[h]
\begin{center}
\small
\setlength{\tabcolsep}{3pt}
\begin{tabular}{lcccc}
\toprule
 & Model 1 & Model 2 & Model 3 & Model 4 \\\midrule
Threshold & 0.000 & 0.750 & 0.800 & 0.918 \\
\% Removed & 0.0\% & 2.5\% & 2.9\% & 5.0\% \\
Aggregate Score & 12.0\% & 12.1\% & \textbf{12.4\%} & \textbf{12.7\%} \\
\bottomrule
\end{tabular}
\end{center}
\caption{Russian Threshold Analysis}
\label{tab:lang_threshold_ru}
\end{table}

% sw
\begin{table}[h]
\begin{center}
\small
\setlength{\tabcolsep}{3pt}
\begin{tabular}{lcccccccc}
\toprule
 & Model 1 & Model 2 & Model 3 & Model 4 & Model 5 & Model 6 & Model 7 & Model 8 \\\midrule
Threshold & 0.000 & 0.075 & 0.098 & 0.132 & 0.167 & 0.186 & 0.300 & 0.544 \\
\% Removed & 0.0\% & 5.0\% & 10.0\% & 20.0\% & 30.0\% & 50.0\% & 64.2\% & 70.0\% \\
Aggregate Score & 8.8\% & 9.7\% & 8.7\% & 8.8\% & 9.2\% & \textbf{10.9\%} & \textbf{10.9\%} & \textbf{11.6\%} \\
\bottomrule
\end{tabular}
\end{center}
\caption{Swahili Threshold Analysis}
\label{tab:lang_threshold_sw}
\end{table}

% te
\begin{table}[h]
\begin{center}
\small
\setlength{\tabcolsep}{3pt}
\begin{tabular}{lcccccccc}
\toprule
 & Model 1 & Model 2 & Model 3 & Model 4 & Model 5 & Model 6 & Model 7 & Model 8 \\\midrule
Threshold & 0.000 & 0.207 & 0.262 & 0.297 & 0.515 & 0.600 & 0.701 & 0.996 \\
\% Removed & 0.0\% & 5.0\% & 10.0\% & 15.0\% & 20.0\% & 22.4\% & 25.0\% & 30.0\% \\
Aggregate Score & 5.4\% & 5.0\% & 5.4\% & 5.1\% & 5.2\% & 5.3\% & \textbf{5.9}\% & 5.4\%\\
\bottomrule
\end{tabular}
\end{center}
\caption{Telugu Threshold Analysis}
\label{tab:lang_threshold_te}
\end{table}

% th
\begin{table}[h]
\begin{center}
\small
\setlength{\tabcolsep}{3pt}
\begin{tabular}{lcccc}
\toprule
 & Model 1 & Model 2 & Model 3 & Model 4 \\\midrule
Threshold & 0.000 & 0.800 & 0.900 & 0.961 \\
\% Removed & 0.0\% & 2.7\% & 3.5\% & 5.0\% \\
Aggregate Score & 6.8\% & 6.2\% & \textbf{6.8\%} & \textbf{6.9\%} \\
\bottomrule
\end{tabular}
\end{center}
\caption{Thai Threshold Analysis}
\label{tab:lang_threshold_th}
\end{table}

% tr
\begin{table}[h]
\begin{center}
\small
\setlength{\tabcolsep}{3pt}
\begin{tabular}{lccccccc}
\toprule
 & Model 1 & Model 2 & Model 3 & Model 4 & Model 5 & Model 6 & Model 7 \\\midrule
Threshold & 0.000 & 0.704 & 0.724 & 0.750 & 0.800 & 0.866 & 0.932 \\
\% Removed & 0.0\% & 5.0\% & 6.1\% & 6.6\% & 7.7\% & 10.0\% & 13.9\% \\
Aggregate Score & 10.9\% & 10.3\% & 9.3\% & 10.2\% & 10.5\% & \textbf{11.5\%} & \textbf{11.4\%} \\
\bottomrule
\end{tabular}
\end{center}
\caption{Turkish Threshold Analysis}
\label{tab:lang_threshold_tr}
\end{table}

\clearpage
% \subsection{Deduplication}

% \begin{figure}[h]
%     \centering \includegraphics[width=0.9\linewidth]{plots/dedup_plot.pdf}
%     \caption{\textbf{Impact of Deduplication} Global deduplication has different impact per language: while providing significant boosts in Russian and Turkish, in Chinese and Arabic the performance uplift is smaller, and for French and Thai it does not improve performance at all. All models represented were trained for a total of 350B tokens on at most 2 epochs, and Hindi, Swahili and Telugu were omitted as they would require a large number of epochs to train at this scale.}
%     \label{fig:dedup_impact}
% \end{figure}

\clearpage
\subsection{Filtering}
\subsubsection{Stopwords}
\label{app:stopwords}

As mentioned in~\cref{stopwords}, we analyzed word frequencies
in our reference datasets (Wikipedia) using our word tokenizers to
identify the most frequently occurring words.

After counting word occurrences directly on the raw data of our reference
datasets, we noticed that some stopwords were actually non-alphabetic
symbols or numbers rather than meaningful words. To refine the list, we
removed all numbers and symbols. If fewer than eight stopwords (eight being the number of stopwords in the original Gopher English stopword list~\citep{rae2022scalinglanguagemodelsmethods}) remained
after this filtering, we lowered the frequency threshold to increase the
number of stopwords and ensure sufficient stopword coverage.

When analyzing English stopwords from the Wikipedia
reference dataset, we found that the original Gopher quality filter did
not necessarily select the most frequent words. This suggested a
different selection criterion had been used. However, since our method
is scalable across languages and performs well in experiments, we
adopted it as our approach and collected stopwords for each language
supported by GlotLID (on Wikipedia when available, and on GlotLID-Corpus
for languages that do not have their own Wikipedia).

When reviewing the languages with the largest amount of data after LID and stopwords filtering, we noticed that some low resource languages had an unexpectedly large amount of data.
For example, Dagbani, a language from the Niger--Congo family with around 1 million
native speakers and low internet presence, ended up with a large amount (2TB) of text data after language
filtering. Through manual inspection, we found that most of this
data was misclassified English and German. We had expected that the stopwords
filter would remove most of this non-Dagbani content; however, the filter removed very little.
Inspecting the list of Dagbani stopwords revealed a high amount of English words
(shown in bold):

\begin{center}
\textbf{the}, ni, \textbf{of}, \textbf{a}, \textbf{in}, ka,
\textbf{and}, o, \textbf{be}, daa, \textbf{to}, di, n, ny\textepsilon la,
\textbf{or}, \textbf{is}
\end{center}

Through further investigation, we found that many other languages had English
stopwords in their list. We traced this issue to the Wikipedias for lower
resource languages, where many articles are directly copied untranslated from
the English wikipedia (for later translation) and some boilerplate/meta
pages exist in the original English. As language classifiers are often
trained on Wikipedia, this may explain why English data is mislabeled as these lowe-resource languages in
the first place.

We ``cleaned'' Wikipedias by: a) removing the notes and references
sections, which sometimes are in other languages and follow a very
specific format; b) dropping articles where the most common script
doesn't match what we expected for the language; c) dropping articles
where our language classifier predicted English with above 70\%
confidence.\\
We then recomputed stopwords on this new clean version of Wikipedia, which resulted in a \textgreater99\% removal rate when filtering Dagbani data using the updated stopwords:

\begin{center}
ni, ka, o, daa, di, n, ny\textepsilon la, din, ti, b\textepsilon, be, ny\textepsilon, maa
\end{center}

\clearpage
\subsubsection{Filtering thresholds}

\paragraph{Filtering details} We employ the following filters from FineWeb with fixed thresholds for all languages, only changing the way ``words'' are defined depending on each language's word-level tokenizer (\cref{sec:separating-words}):

\begin{itemize}
    \item \textbf{FineWeb Quality filters}: ratio of characters in duplicate lines $\leq$ 0.1;
    \item \textbf{Gopher Quality filters}: 50 $\leq$ \#words $\leq$ 100000; ratio of symbols to \#words $\leq$ 0.1; ratio of bullet points to \#lines $\leq$ 0.9; ratio of ellipsis to \#lines $\leq$ 0.3, stop words in document $\geq$ 2 (with stopwords determined following~\cref{stopwords});
\end{itemize}

We \textit{tune} the following filters with the different adaptation methods we consider:
\begin{itemize}
    \item \textbf{FineWeb Quality filters}: maximum ratio of lines not ending with punctuation; maximum ratio of \#lines to \#words
    \item \textbf{Gopher Quality filters}: maximum average word length; minimum average word length; maximum ratio of non alphabetic words;
    \item \textbf{Gopher Repetition filters}: fraction of duplicate lines, fraction of characters taken up by the most common 2-, 3- and 4-grams; fraction of characters taken up by every single repeated 5-, 6-, 7-, 8-, 9-, and 10-gram 
\end{itemize}

Results from training models on data obtained by applying the different adaptations methods to each group filters can be seen in~\cref{tab:filtering-runs-table}. We \textbf{select the best performing method for each filter group} (marked in bold) for our pipeline. We also show the average removal rates across languages of each method in~\cref{tab:filtering-percentages}.

\begin{table}[h]
\small
\setlength{\tabcolsep}{3pt}
\centering
\begin{tabular}{l|cc|cccc|cccc}
\toprule
  Filter  & & & \multicolumn{4}{c}{cc} & \multicolumn{4}{|c}{wiki} \\
 Group & Baseline & English & 10Tail & MeanStd & MedRatio & Quant & 10Tail & MeanStd & MedRatio & Quant \\
\midrule
\textbf{fwq} & 7.00 & - & - & 5.22 & 4.00 & 4.33 & \textbf{3.00} & 5.00 & 3.89 & 3.56 \\
\textbf{goq} & 6.33 & - & 5.22 & - & 3.89 & 4.56 & 4.44 & 4.11 & 4.22 & \textbf{3.22} \\
\textbf{gor} & 6.22 & 4.22 & 3.33 & \textbf{2.22} & - & 4.11 & - & 3.89 & - & 4.00 \\
\bottomrule
\end{tabular}
\caption{\textbf{Average ranks} by block and method across all languages. \textit{Baseline} has no filtering, English is the default FineWeb English thresholds. We then compute each of the other 4 methods -- 10Tail, MeanStd, MedRatio (MedianRatio), and Quantile (Quant) --  on both Common Crawl (cc) data and on Wikipedia (wiki). Cells marked with - correspond to method-filter-group combinations that would remove over 75\% of data with a single filter on at least one of the languages, or that would not remove anything at all. Lower ranks are better.}
\label{tab:filtering-runs-table}
\end{table}

\begin{table}[h]
\small
\setlength{\tabcolsep}{3pt}
\centering
\begin{tabular}{l|c|cccc|cccc}
\toprule
  Filter  & & \multicolumn{4}{c}{cc} & \multicolumn{4}{|c}{wiki} \\
 Group & English & 10Tail & MeanStd & MedRatio & Quant & 10Tail & MeanStd & MedRatio & Quant \\
\midrule
\textbf{fwq} & - & - & 36.81\% & 38.03\% & 33.82\% & 40.35\% & 38.04\% & 37.61\% & 44.31\% \\
\textbf{goq} & - & 41.42\% & - & 47.09\% & 45.14\% & 49.23\% & 47.58\% & 46.90\% & 46.81\% \\
\textbf{gor} & 26.39\% & 29.53\% & 25.63\% & - & 26.08\% & - & 24.79\% & - & 26.50\% \\
\bottomrule
\end{tabular}
\caption{\textbf{Average removal rates} by method across datasets. Values represent percentage of data filtered.}
\label{tab:filtering-percentages}
\end{table}

\clearpage
\subsubsection{Precision filtering lower resource languages}
\label{app:precision-lower-resource}

Language Identification precision computed on a balanced test set does not correspond to the precision on web crawled data, due to class imbalance between high- and low-resource languages.
Precision on the crawled corpora can be calculated as in \citet{caswell-etal-2020-language}, where $x$ is the real proportion of the target language in the full web crawl:

\[
\text{precision}_{\text{crawl}} = \frac{x \cdot \text{recall}}{x \cdot \text{recall} + (1 - x) \cdot \text{fpr}}
\]

For low-resource languages (low \( x \)), a low false positive rate (\( \text{fpr} \)) is crucial, as higher false positives significantly reduce precision. For high-resource languages, web presence is high, so false positives are less critical.

If a low-resource language is sufficiently distinct from high-resource languages, the false positive rate will often be low. However, if a closely related high-resource language exists, the high-resource language may be misclassified as low-resource. In such scenarios, n-gram-based LID fails because common n-grams lead to misclassification of high-resource language sentences as low-resource. 
\paragraph{Wordlist filtering} To maintain high precision for low-resource languages after LID, wordlist filtering is suggested to retain in-language documents~\citep{caswell-etal-2020-language, bapna2022buildingmachinetranslationsystems}.   

To build such wordlists, we propose a simple approach: only consider tokens whose affinity exceeds a high threshold (we use \( \gamma = 0.85 \)) for each language. The affinity of a token \( t \) in language \( l \) is defined as:

\[
\text{Affinity}(t, l) = \frac{f_{t,l}}{\sum_{l' \in \mathcal{L}} f_{t,l'}}
\]

where \( f_{t,l} \) is the raw count of token \( t \) in language \( l \), and \( \sum_{l' \in \mathcal{L}} f_{t,l'} \) is the total count of token \( t \) across all languages in the set \( \mathcal{L} \). A text labeled as a low-resource language \( l \) is considered in-language if some of its words appear in the wordlist created for \( l \); otherwise, it is considered contaminated. We used data from the GlotLID-Corpus~\citep{kargaran-etal-2023-glotlid} to create wordlists, applying the tokenizer specific to each language from \cref{app:word-tokenizers}. For each language $l$, the same tokenizer (the tokenizer of language $l$) is used to compute $f_{t,l'}$ to ensure consistent separation of words.

\begin{figure}[h]
    \centering \includegraphics[width=0.9\linewidth]{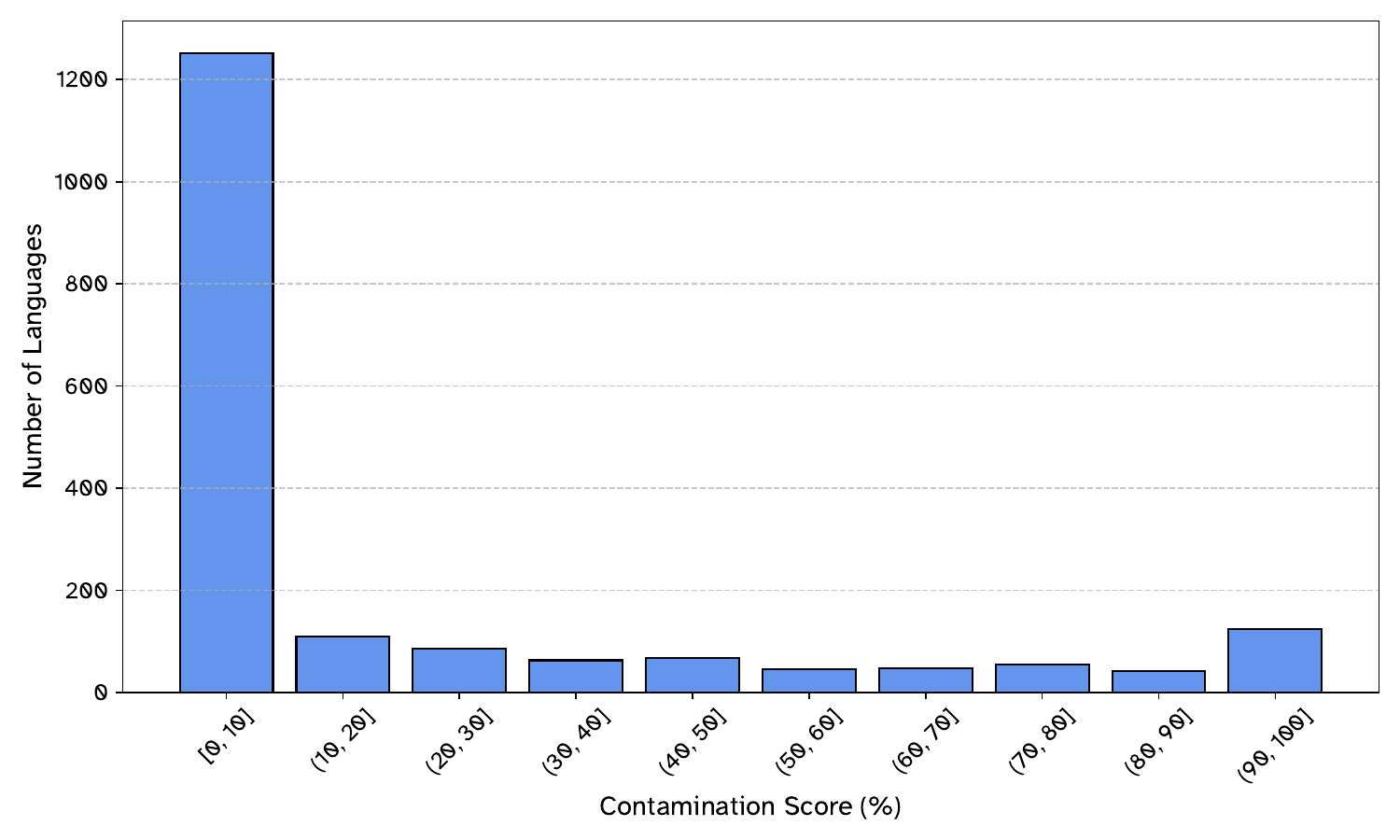}
    \caption{\textbf{Contamination scores for 1,900 languages via wordlist filtering.} The plot indicates that the majority of the languages have their data in-language (non-contaminated).}
    \label{fig:contamination_score}
\end{figure}

We use wordlist filtering as an indicator of contamination, where the contamination score is defined as the percentage of documents removed by the filter. This helps identify languages with low quality for manual auditing. We select 10,000 random documents from each language and calculate the percentage of documents filtered. \texttt{glk\_Arab} is one of the languages with the highest contamination score. We present the distribution of contamination scores for 1,900 languages for which we have wordlists in~\cref{fig:contamination_score}. The majority of the languages have their data in-language (non-contaminated). However, around a third of them have contamination scores above 10\%.

\paragraph{URL Whitelist} Manual inspection of the filtering process revealed that some of the wordlists were too strict. This was the case of some English-based Pidgin languages, such as Nigerian Pidgin, for example, where the resulting wordlist was relatively short. To avoid excessive filtering caused by strict wordlists, we additionally kept documents removed by the wordlist filtering whose URLs contained specific terms related to the language (the language code, the name of the language, possibly top level domains for that region and/or names of regions where the language is spoken). For Nigerian Pidgin, this list contained the following words: ``pcm'', ``pidgin'', ``naija'', ``.ng'', ``nigerianpidgin'', ``nigeria'', and ``nigerian''. We show example URLs containing in-language content that had been removed by wordlist filtering that are then caught by the URL Whitelist in~\cref{tab:nigerian-pidgin-examples}.

\begin{table}[h]
\begin{center}
\small
\setlength{\tabcolsep}{3pt}
\begin{tabular}{p{10cm}l}
\toprule
\textbf{URL} & \textbf{Matched Words} \\\midrule
\url{http://www.supersport.com/football/nigeria-naija/news/121221/Uefa_don_ban_Malaga} & nigeria, nigerian, naija \\
\url{https://manutdinpidgin.com/2018/06/28/manchester-united-target-sergej-milinkovic-savic-don-react-ontop-the-transfer-rumour/} & pidgin \\
\url{https://pcm.wikipedia.org/wiki/Japan} & pcm \\
\url{https://www.bbc.com/pidgin/sport-43612518} & pidgin \\
\bottomrule
\end{tabular}
\end{center}
\caption{Matched words for selected URLs in Nigerian Pidgin}
\label{tab:nigerian-pidgin-examples}
\end{table}

\paragraph{Filtering results}
We audit three low-resource languages: \text{glk\_Arab}, \text{bar\_Latn}, and \text{ary\_Arab}, by asking native speakers to manually label 2,000 randomly sampled documents as being in-language or not. The results of applying wordlist filtering with the URL Whitelist to these languages are shown in \cref{tab:wordlist-audit}. Applying wordlist filtering maintains recall while improving precision for both glk\_Arab and bar\_Latn. However, for ary\_Arab, the improvement is not very significant. This is because the training data for LID does not adequately represent ary\_Arab. Precision could be increased further by requiring a certain fraction of the document to be contained in the wordlist (instead of just a single word), but this would require manual tuning and could result in a drop in recall.

%thresholding and retaining only those documents where more than 3\% of the tokens belong to the wordlist, precision improves. For the other languages, precision can also be further improved through thresholding, especially with larger values for the percentage of tokens in the wordlist, although this may result in a loss of recall and requires annotations to tune.

\begin{table}[h]
\begin{center}

\small
\setlength{\tabcolsep}{4pt}
\begin{tabular}{l|c|cc}
\toprule
 & \multicolumn{1}{c|}{Pre-filtering} & \multicolumn{2}{c}{Filtering} \\
Language & Precision & Recall & Precision \\
\midrule
glk\_Arab & 2.10\% & 95.24\% & 27.21\%   \\
bar\_Latn & 69.45\% & 97.77\% & 94.90\%  \\
ary\_Arab & 1.75\% & 88.57\% & 4.14\%   \\
\bottomrule
\end{tabular}
\caption{Evaluation Results for wordlist filtering Based on the Audit}
\label{tab:wordlist-audit}
\end{center}
\end{table}

We publicly release our wordlists and code.\footnote{\url{https://github.com/huggingface/fineweb-2/tree/main/misc/precision_filtering}}

% \begin{table}[h]
% \begin{center}

% \small
% \setlength{\tabcolsep}{4pt}
% \begin{tabular}{l|c||cc|cc}
% \toprule
%  & \multicolumn{1}{c||}{No Filtering} & \multicolumn{2}{c|}{Filtering} & \multicolumn{2}{c}{3\% Thresholded Filtering} \\
% Language & Precision & Recall & Precision & Recall & Precision \\
% \midrule
% glk\_Arab & 2.10\% & 95.24\% & 66.67\% & 64.29\% & 100.00\%  \\
% bar\_Latn & 69.45\% & 96.83\% & 95.39\% & 81.86\% & 97.85\%  \\
% ary\_Arab & 1.75\% & 80.00\% & 5.12\%  & 51.43\% & 22.50\%  \\
% \bottomrule
% \end{tabular}
% \caption{Evaluation Results for wordlist filtering Based on the Audit}
% \label{tab:wordlist-audit}
% \end{center}
% \end{table}

% \clearpage
% \subsection{Deduplication and Rehydration}
% \begin{figure}[h]
%     \centering \includegraphics[width=0.9\linewidth]{plots/removal_rate_by_cluster_size.pdf}
%     \caption{\textbf{Filtering rates by MinHash cluster size} for French documents. The global filtering rate represents the overall percentage of documents removed during the full filtering process. Individual filtering rates are shown for each cluster size, providing a proxy for cluster quality—higher removal rates may indicate lower-quality clusters.}
%     \label{fig:rehydration_removal_rates}
% \end{figure}

% \begin{figure}[h]
%     \centering \includegraphics[width=0.9\linewidth]{plots/cluster_weights.pdf}
%     \caption{\textbf{Upsampling weight} assigned to each MinHash cluster size for French documents.}
%     \label{fig:rehydration_upsampling_weights}
% \end{figure}

\clearpage
\subsection{Improvement from each pipeline step}
\begin{itemize}
    \item \textbf{LID} Language Identification and threshold
    \item \textbf{LID + D} LID \& global MinHash deduplication
    \item \textbf{LID + D + F} LID + D \& heuristic filtering
    \item \textbf{FW2 (R)} LID + D + F \& rehydration (deduplication informed upsampling)
\end{itemize}

% Arabic Table
\begin{table}[h]
\begin{center}
\label{tab:results_ar}
\small
\setlength{\tabcolsep}{3pt}
\begin{tabular}{p{3cm}|c||cc|cc|cc|cc}
\toprule
 & \multicolumn{1}{c||}{Random} & \multicolumn{2}{c|}{LID} & \multicolumn{2}{c|}{LID + D} & \multicolumn{2}{c|}{LID + D + F} & \multicolumn{2}{c}{FW2 (R)} \\
Task & Baseline & Raw & Rescaled & Raw & Rescaled & Raw & Rescaled & Raw & Rescaled \\
\midrule
Alghafa: MCQ Exams (GK) & 25.0 & \textbf{37.1} & \textbf{16.1} & 35.5 & 14.0 & 35.6 & 14.1 & 36.3 & 15.1 \\
Belebele (RC) & 25.0 & 32.1 & 9.4 & 31.4 & 8.5 & 32.6 & 10.1 & \textbf{33.9} & \textbf{11.9} \\
Alghafa: SOQAL (RC) & 20.0 & 64.0 & 55.0 & 62.3 & 52.9 & \textbf{67.9} & \textbf{59.9} & \textbf{67.9} & \textbf{59.9} \\
Alghafa: ARC Easy (GK) & 25.0 & 38.1 & 17.5 & 39.2 & 18.9 & 40.9 & 21.2 & \textbf{41.1} & \textbf{21.5} \\
Okapi: Hellaswag (NLU) & 25.0 & 36.4 & 15.2 & 39.0 & 18.6 & 40.9 & 21.2 & \textbf{43.3} & \textbf{24.4} \\
OALL2024: PIQA (RES) & 50.0 & 58.4 & 16.8 & 60.9 & 21.8 & 61.8 & 23.6 & \textbf{61.9} & \textbf{23.8} \\
OALL2024: RACE (RC) & 25.0 & 32.5 & 9.9 & 32.9 & 10.5 & 33.9 & 11.8 & \textbf{34.4} & \textbf{12.5} \\
OALL2024: SCIQ (GK) & 25.0 & 66.8 & 55.7 & 67.3 & 56.4 & \textbf{68.8} & \textbf{58.4} & 67.9 & 57.2 \\
X-CODAH (RES) & 25.0 & 35.8 & 14.4 & 34.0 & 12.0 & \textbf{40.0} & \textbf{19.9} & 38.5 & 18.0 \\
X-CSQA (RES) & 20.0 & 32.8 & 16.0 & 32.5 & 15.6 & 32.7 & 15.8 & \textbf{34.2} & \textbf{17.8} \\
X-Story Cloze (NLU) & 50.0 & 59.0 & 18.1 & 58.9 & 17.8 & 59.8 & 19.7 & \textbf{60.9} & \textbf{21.9} \\
ARCD (GK) & 0.0 & 29.9 & 29.9 & 32.0 & 32.0 & 33.0 & 33.0 & \textbf{33.1} & \textbf{33.1} \\
MLQA (RC) & 0.0 & 22.2 & 22.2 & 21.5 & 21.5 & 21.3 & 21.3 & \textbf{23.2} & \textbf{23.2} \\
Tydiqa (RC) & 0.0 & \textbf{39.5} & \textbf{39.5} & 37.9 & 37.9 & 36.8 & 36.8 & 36.5 & 36.5 \\
ArabicMMLU (GK) & 28.0 & 39.9 & 16.6 & 40.0 & 16.7 & 40.1 & 16.9 & \textbf{41.1} & \textbf{18.2} \\
\midrule
GK tasks & - & \multicolumn{2}{c|}{27.2} & \multicolumn{2}{c|}{27.6} & \multicolumn{2}{c|}{28.7} & \multicolumn{2}{c}{\textbf{29.0}} \\
RC tasks & - & \multicolumn{2}{c|}{27.2} & \multicolumn{2}{c|}{26.2} & \multicolumn{2}{c|}{28.0} & \multicolumn{2}{c}{\textbf{28.8}} \\
RES tasks & - & \multicolumn{2}{c|}{15.7} & \multicolumn{2}{c|}{16.5} & \multicolumn{2}{c|}{19.8} & \multicolumn{2}{c}{\textbf{19.8}} \\
NLU tasks & - & \multicolumn{2}{c|}{16.6} & \multicolumn{2}{c|}{18.2} & \multicolumn{2}{c|}{20.4} & \multicolumn{2}{c}{\textbf{23.1}} \\
\midrule
Aggregate Score & - & \multicolumn{2}{c|}{21.7} & \multicolumn{2}{c|}{22.1} & \multicolumn{2}{c|}{24.2} & \multicolumn{2}{c}{\textbf{25.2}} \\
\bottomrule
\end{tabular}
\caption{Arabic Results}
\end{center}
\end{table}

% French Table
\begin{table}[h]
\begin{center}
\label{tab:results_fr}
\small
\setlength{\tabcolsep}{3pt}
\begin{tabular}{p{3cm}|c||cc|cc|cc|cc}
\toprule
 & \multicolumn{1}{c||}{Random} & \multicolumn{2}{c|}{LID} & \multicolumn{2}{c|}{LID + D} & \multicolumn{2}{c|}{LID + D + F} & \multicolumn{2}{c}{FW2 (R)} \\
Task & Baseline & Raw & Rescaled & Raw & Rescaled & Raw & Rescaled & Raw & Rescaled \\
\midrule
Okapi: ARC (GK) & 25.0 & 31.3 & 8.4 & 30.0 & 6.7 & 31.9 & 9.2 & \textbf{33.0} & \textbf{10.6} \\
Belebele (RC) & 25.0 & 33.8 & 11.7 & 33.1 & 10.8 & 35.2 & 13.5 & \textbf{36.0} & \textbf{14.6} \\
Okapi: Hellaswag (NLU) & 25.0 & 45.1 & 26.8 & 47.3 & 29.7 & 51.8 & 35.8 & \textbf{52.6} & \textbf{36.8} \\
X-CODAH (RES) & 25.0 & 37.0 & 15.9 & 37.0 & 16.0 & 40.5 & 20.6 & \textbf{42.3} & \textbf{23.1} \\
X-CSQA (RES) & 20.0 & 38.0 & 22.5 & 34.6 & 18.2 & 39.8 & 24.7 & \textbf{40.4} & \textbf{25.5} \\
FQuad (RC) & 0.0 & 28.0 & 28.0 & 27.5 & 27.5 & 29.3 & 29.3 & \textbf{35.0} & \textbf{35.0} \\
Mintaka (GK) & 0.0 & \textbf{9.5} & \textbf{9.5} & 7.5 & 7.5 & 8.9 & 8.9 & 8.1 & 8.1 \\
Meta MMLU (GK) & 25.0 & 28.3 & 4.4 & 28.4 & 4.5 & 29.0 & 5.3 & \textbf{29.6} & \textbf{6.1} \\
\midrule
GK tasks & - & \multicolumn{2}{c|}{7.4} & \multicolumn{2}{c|}{6.2} & \multicolumn{2}{c|}{7.8} & \multicolumn{2}{c}{\textbf{8.3}} \\
RC tasks & - & \multicolumn{2}{c|}{19.9} & \multicolumn{2}{c|}{19.2} & \multicolumn{2}{c|}{21.4} & \multicolumn{2}{c}{\textbf{24.8}} \\
RES tasks & - & \multicolumn{2}{c|}{19.2} & \multicolumn{2}{c|}{17.1} & \multicolumn{2}{c|}{22.7} & \multicolumn{2}{c}{\textbf{24.3}} \\
NLU tasks & - & \multicolumn{2}{c|}{26.8} & \multicolumn{2}{c|}{29.7} & \multicolumn{2}{c|}{35.8} & \multicolumn{2}{c}{\textbf{36.8}} \\
\midrule
Aggregate Score & - & \multicolumn{2}{c|}{18.3} & \multicolumn{2}{c|}{18.0} & \multicolumn{2}{c|}{21.9} & \multicolumn{2}{c}{\textbf{23.6}} \\
\bottomrule
\end{tabular}
\caption{French Results}
\end{center}
\end{table}

% Russian Table
\begin{table}[h]
\begin{center}
\label{tab:results_ru}
\small
\setlength{\tabcolsep}{3pt}
\begin{tabular}{p{3cm}|c||cc|cc|cc|cc}
\toprule
 & \multicolumn{1}{c||}{Random} & \multicolumn{2}{c|}{LID} & \multicolumn{2}{c|}{LID + D} & \multicolumn{2}{c|}{LID + D + F} & \multicolumn{2}{c}{FW2 (R)} \\
Task & Baseline & Raw & Rescaled & Raw & Rescaled & Raw & Rescaled & Raw & Rescaled \\
\midrule
Okapi: ARC (GK) & 25.0 & 29.1 & 5.4 & 30.4 & 7.1 & \textbf{33.8} & \textbf{11.7} & 32.2 & 9.6 \\
Belebele (RC) & 25.0 & 34.0 & 12.0 & 33.7 & 11.6 & 34.8 & 13.0 & \textbf{36.4} & \textbf{15.2} \\
Okapi: Hellaswag (NLU) & 25.0 & 41.0 & 21.3 & 43.6 & 24.8 & 45.8 & 27.8 & \textbf{46.8} & \textbf{29.0} \\
Parus (RES) & 50.0 & 64.9 & 29.9 & 65.7 & 31.4 & \textbf{68.2} & \textbf{36.4} & 68.1 & 36.2 \\
OpenBookQA (RES) & 25.0 & 36.0 & 14.7 & 36.0 & 14.7 & 35.9 & 14.5 & \textbf{38.3} & \textbf{17.7} \\
X-CODAH (RES) & 25.0 & 33.9 & 11.8 & 34.9 & 13.2 & 35.4 & 13.8 & \textbf{37.1} & \textbf{16.2} \\
X-CSQA (RES) & 20.0 & 35.3 & 19.2 & 37.4 & 21.8 & 35.0 & 18.7 & \textbf{38.6} & \textbf{23.3} \\
X-Story Cloze (NLU) & 50.0 & 66.9 & 33.7 & 66.7 & 33.5 & 68.7 & 37.5 & \textbf{69.4} & \textbf{38.9} \\
Sber SQuAD (RC) & 0.0 & 27.8 & 27.8 & 32.4 & 32.4 & 32.9 & 32.9 & \textbf{37.1} & \textbf{37.1} \\
Tydiqa (RC) & 0.0 & 29.9 & 29.9 & 32.4 & 32.4 & \textbf{36.7} & \textbf{36.7} & 35.5 & 35.5 \\
X-QuAD (RC) & 0.0 & 19.6 & 19.6 & 22.8 & 22.8 & 23.6 & 23.6 & \textbf{25.2} & \textbf{25.2} \\
RUMMLU (GK) & 25.0 & 29.3 & 5.7 & 29.0 & 5.4 & 29.7 & 6.3 & \textbf{30.1} & \textbf{6.8} \\
\midrule
GK tasks & - & \multicolumn{2}{c|}{5.6} & \multicolumn{2}{c|}{6.3} & \multicolumn{2}{c|}{\textbf{9.0}} & \multicolumn{2}{c}{8.2} \\
RC tasks & - & \multicolumn{2}{c|}{22.3} & \multicolumn{2}{c|}{24.8} & \multicolumn{2}{c|}{26.5} & \multicolumn{2}{c}{\textbf{28.2}} \\
RES tasks & - & \multicolumn{2}{c|}{18.9} & \multicolumn{2}{c|}{20.2} & \multicolumn{2}{c|}{20.9} & \multicolumn{2}{c}{\textbf{23.4}} \\
NLU tasks & - & \multicolumn{2}{c|}{27.5} & \multicolumn{2}{c|}{29.1} & \multicolumn{2}{c|}{32.6} & \multicolumn{2}{c}{\textbf{34.0}} \\
\midrule
Aggregate Score & - & \multicolumn{2}{c|}{18.6} & \multicolumn{2}{c|}{20.1} & \multicolumn{2}{c|}{22.3} & \multicolumn{2}{c}{\textbf{23.4}} \\
\bottomrule
\end{tabular}
\caption{Russian Results}
\end{center}
\end{table}

% Thai Table
\begin{table}[h]
\begin{center}
\label{tab:results_th}
\small
\setlength{\tabcolsep}{3pt}
\begin{tabular}{p{3cm}|c||cc|cc|cc|cc}
\toprule
 & \multicolumn{1}{c||}{Random} & \multicolumn{2}{c|}{LID} & \multicolumn{2}{c|}{LID + D} & \multicolumn{2}{c|}{LID + D + F} & \multicolumn{2}{c}{FW2 (R)} \\
Task & Baseline & Raw & Rescaled & Raw & Rescaled & Raw & Rescaled & Raw & Rescaled \\
\midrule
Belebele (RC) & 25.0 & 31.6 & 8.7 & 31.5 & 8.7 & 32.0 & 9.4 & \textbf{32.9} & \textbf{10.5} \\
Translated Hellaswag (NLU) & 25.0 & 32.5 & 10.0 & 33.1 & 10.8 & 35.9 & 14.5 & \textbf{35.9} & \textbf{14.5} \\
M3Exam (GK) & 22.9 & 27.6 & 6.1 & \textbf{28.1} & \textbf{6.7} & 27.5 & 5.9 & 28.1 & 6.7 \\
ThaiQA (RC) & 0.0 & \textbf{27.2} & \textbf{27.2} & 23.8 & 23.8 & 22.1 & 22.1 & 26.3 & 26.3 \\
X-QuAD (RC) & 0.0 & 19.6 & 19.6 & 18.6 & 18.6 & 17.3 & 17.3 & \textbf{20.8} & \textbf{20.8} \\
Meta MMLU (GK) & 25.0 & 27.6 & 3.4 & 27.4 & 3.2 & 28.1 & 4.2 & \textbf{28.4} & \textbf{4.6} \\
\midrule
GK tasks & - & \multicolumn{2}{c|}{4.7} & \multicolumn{2}{c|}{5.0} & \multicolumn{2}{c|}{5.1} & \multicolumn{2}{c}{\textbf{5.6}} \\
RC tasks & - & \multicolumn{2}{c|}{18.5} & \multicolumn{2}{c|}{17.0} & \multicolumn{2}{c|}{16.2} & \multicolumn{2}{c}{\textbf{19.2}} \\
NLU tasks & - & \multicolumn{2}{c|}{10.0} & \multicolumn{2}{c|}{10.8} & \multicolumn{2}{c|}{14.5} & \multicolumn{2}{c}{\textbf{14.5}} \\
\midrule
Aggregate Score & - & \multicolumn{2}{c|}{11.1} & \multicolumn{2}{c|}{11.0} & \multicolumn{2}{c|}{11.9} & \multicolumn{2}{c}{\textbf{13.1}} \\
\bottomrule
\end{tabular}
\caption{Thai Results}
\end{center}
\end{table}

% Turkish Table
\begin{table}[h]
\begin{center}
\label{tab:results_tr}
\small
\setlength{\tabcolsep}{3pt}
\begin{tabular}{p{3cm}|c||cc|cc|cc|cc}
\toprule
 & \multicolumn{1}{c||}{Random} & \multicolumn{2}{c|}{LID} & \multicolumn{2}{c|}{LID + D} & \multicolumn{2}{c|}{LID + D + F} & \multicolumn{2}{c}{FW2 (R)} \\
Task & Baseline & Raw & Rescaled & Raw & Rescaled & Raw & Rescaled & Raw & Rescaled \\
\midrule
TR Leaderboard: ARC (GK) & 25.0 & 43.7 & 25.0 & 45.1 & 26.8 & \textbf{47.9} & \textbf{30.6} & 46.3 & 28.4 \\
Belebele (RC) & 25.0 & 31.5 & 8.7 & 32.3 & 9.7 & 33.0 & 10.7 & \textbf{34.2} & \textbf{12.2} \\
Okapi: Hellaswag (NLU) & 25.0 & 42.4 & 23.3 & 43.3 & 24.3 & 45.3 & 27.1 & \textbf{46.8} & \textbf{29.1} \\
X-COPA (RES) & 50.0 & 60.7 & 21.3 & 60.7 & 21.5 & \textbf{62.8} & \textbf{25.6} & 62.7 & 25.3 \\
THQuAD (RC) & 0.0 & 20.4 & 20.4 & 25.6 & 25.6 & 20.6 & 20.6 & \textbf{26.1} & \textbf{26.1} \\
X-QuAD (RC) & 0.0 & 15.8 & 15.8 & 18.2 & 18.2 & 15.1 & 15.1 & \textbf{20.2} & \textbf{20.2} \\
Exams (GK) & 23.4 & 29.4 & 7.8 & 29.3 & 7.7 & 28.8 & 7.1 & \textbf{30.7} & \textbf{9.6} \\
TR Leaderboard: MMLU (GK) & 25.0 & 29.8 & 6.4 & \textbf{30.0} & \textbf{6.7} & 29.8 & 6.5 & 29.2 & 5.7 \\
\midrule
GK tasks & - & \multicolumn{2}{c|}{13.1} & \multicolumn{2}{c|}{13.7} & \multicolumn{2}{c|}{\textbf{14.7}} & \multicolumn{2}{c}{14.6} \\
RC tasks & - & \multicolumn{2}{c|}{15.0} & \multicolumn{2}{c|}{17.8} & \multicolumn{2}{c|}{15.5} & \multicolumn{2}{c}{\textbf{19.5}} \\
RES tasks & - & \multicolumn{2}{c|}{21.3} & \multicolumn{2}{c|}{21.5} & \multicolumn{2}{c|}{\textbf{25.6}} & \multicolumn{2}{c}{25.3} \\
NLU tasks & - & \multicolumn{2}{c|}{23.3} & \multicolumn{2}{c|}{24.3} & \multicolumn{2}{c|}{27.1} & \multicolumn{2}{c}{\textbf{29.1}} \\
\midrule
Aggregate Score & - & \multicolumn{2}{c|}{18.2} & \multicolumn{2}{c|}{19.3} & \multicolumn{2}{c|}{20.7} & \multicolumn{2}{c}{\textbf{22.1}} \\
\bottomrule
\end{tabular}
\caption{Turkish Results}
\end{center}
\end{table}

% Chinese Table
\begin{table}[h]
\begin{center}
\label{tab:results_zh}
\small
\setlength{\tabcolsep}{3pt}
\begin{tabular}{p{3cm}|c||cc|cc|cc|cc}
\toprule
 & \multicolumn{1}{c||}{Random} & \multicolumn{2}{c|}{LID} & \multicolumn{2}{c|}{LID + D} & \multicolumn{2}{c|}{LID + D + F} & \multicolumn{2}{c}{FW2 (R)} \\
Task & Baseline & Raw & Rescaled & Raw & Rescaled & Raw & Rescaled & Raw & Rescaled \\
\midrule
Belebele (RC) & 25.0 & 32.3 & 9.8 & 33.2 & 11.0 & 33.0 & 10.7 & \textbf{34.0} & \textbf{12.0} \\
C3 (RC) & 27.1 & 47.6 & 28.2 & 47.2 & 27.5 & \textbf{50.6} & \textbf{32.2} & 49.2 & 30.3 \\
Okapi: Hellaswag (NLU) & 25.0 & 38.3 & 17.7 & 38.6 & 18.1 & 41.4 & 21.9 & \textbf{42.2} & \textbf{22.9} \\
M3Exam (GK) & 25.9 & 32.8 & 9.3 & 32.6 & 9.0 & 34.1 & 11.1 & \textbf{34.3} & \textbf{11.3} \\
X-CODAH (RES) & 25.0 & 34.0 & 12.0 & 32.6 & 10.2 & 35.3 & 13.7 & \textbf{39.0} & \textbf{18.6} \\
X-CSQA (RES) & 20.0 & 38.9 & 23.6 & \textbf{41.9} & \textbf{27.4} & 41.2 & 26.6 & 39.8 & 24.7 \\
X-COPA (RES) & 50.0 & 60.9 & 21.8 & 62.5 & 25.0 & 62.0 & 24.0 & \textbf{64.5} & \textbf{28.9} \\
X-Story Cloze (NLU) & 50.0 & 63.0 & 26.0 & 61.6 & 23.1 & 63.1 & 26.2 & \textbf{65.5} & \textbf{30.9} \\
X-Winograd (NLU) & 50.0 & 70.2 & 40.3 & 70.9 & 41.7 & 72.1 & 44.2 & \textbf{74.9} & \textbf{49.8} \\
Chinese SQuAD (RC) & 0.0 & 23.5 & 23.5 & 24.1 & 24.1 & 24.1 & 24.1 & \textbf{26.3} & \textbf{26.3} \\
CMRC (RC) & 0.0 & 38.2 & 38.2 & 38.0 & 38.0 & 38.8 & 38.8 & \textbf{40.2} & \textbf{40.2} \\
MLQA (RC) & 0.0 & 26.8 & 26.8 & 27.8 & 27.8 & 28.5 & 28.5 & \textbf{29.5} & \textbf{29.5} \\
AGIEval (ZH subset) (GK) & 26.8 & 32.9 & 8.3 & 33.4 & 9.1 & \textbf{34.1} & \textbf{10.0} & 33.8 & 9.6 \\
C-Eval (GK) & 25.0 & 31.6 & 8.8 & 32.1 & 9.5 & 32.6 & 10.1 & \textbf{32.7} & \textbf{10.3} \\
CMMLU (GK) & 25.0 & 32.0 & 9.4 & 33.0 & 10.7 & 34.1 & 12.2 & \textbf{34.3} & \textbf{12.4} \\
\midrule
GK tasks & - & \multicolumn{2}{c|}{8.9} & \multicolumn{2}{c|}{9.6} & \multicolumn{2}{c|}{10.9} & \multicolumn{2}{c}{\textbf{10.9}} \\
RC tasks & - & \multicolumn{2}{c|}{25.3} & \multicolumn{2}{c|}{25.7} & \multicolumn{2}{c|}{26.9} & \multicolumn{2}{c}{\textbf{27.7}} \\
RES tasks & - & \multicolumn{2}{c|}{19.1} & \multicolumn{2}{c|}{20.9} & \multicolumn{2}{c|}{21.4} & \multicolumn{2}{c}{\textbf{24.1}} \\
NLU tasks & - & \multicolumn{2}{c|}{28.0} & \multicolumn{2}{c|}{27.7} & \multicolumn{2}{c|}{30.8} & \multicolumn{2}{c}{\textbf{34.6}} \\
\midrule
Aggregate Score & - & \multicolumn{2}{c|}{20.3} & \multicolumn{2}{c|}{20.9} & \multicolumn{2}{c|}{22.5} & \multicolumn{2}{c}{\textbf{24.3}} \\
\bottomrule
\end{tabular}
\caption{Chinese Results}
\end{center}
\end{table}

\clearpage
\subsection{Dataset comparison on Canary Languages}
\label{app:canaray-results}
In addition to the Reference datasets (\cref{sec:reference-datasets}, we compare FineWeb2 with the concurrent work \citet{degibert2024newmassivemultilingualdataset}, as well as with the following language-specific datasets:
\begin{itemize}
    \item \textbf{Arabic}: ArabicWeb24~\citep{farhat2024arabicweb24}, Arabic-101B~\citep{aloui2024101}
    \item \textbf{French}: Croissant~\citep{faysse2024croissantllm}
    \item \textbf{Hindi \& Telugu}: Sangraha~\citep{khan2024indicllmsuite}
    \item \textbf{Hindi}: Odaigen~\citep{parida2024hindi}
    \item \textbf{Russian}: Omnia Russica~\citep{omnia2024}
    \item \textbf{Thai}: Sea CommonCrawl~\citep{dou2025sailor2}
    \item \textbf{Turkish}: VNGRS-Web-Corpus~\citep{turker2024vbart}
    \item \textbf{Chinese}: MNBVC~\citep{mnbvc2023}, TigerBot~\citep{tigerbot2023}, MAP-CC~\citep{du2024chinese}
\end{itemize}

\begin{figure}[h]
\centering
\includegraphics[width=\textwidth]{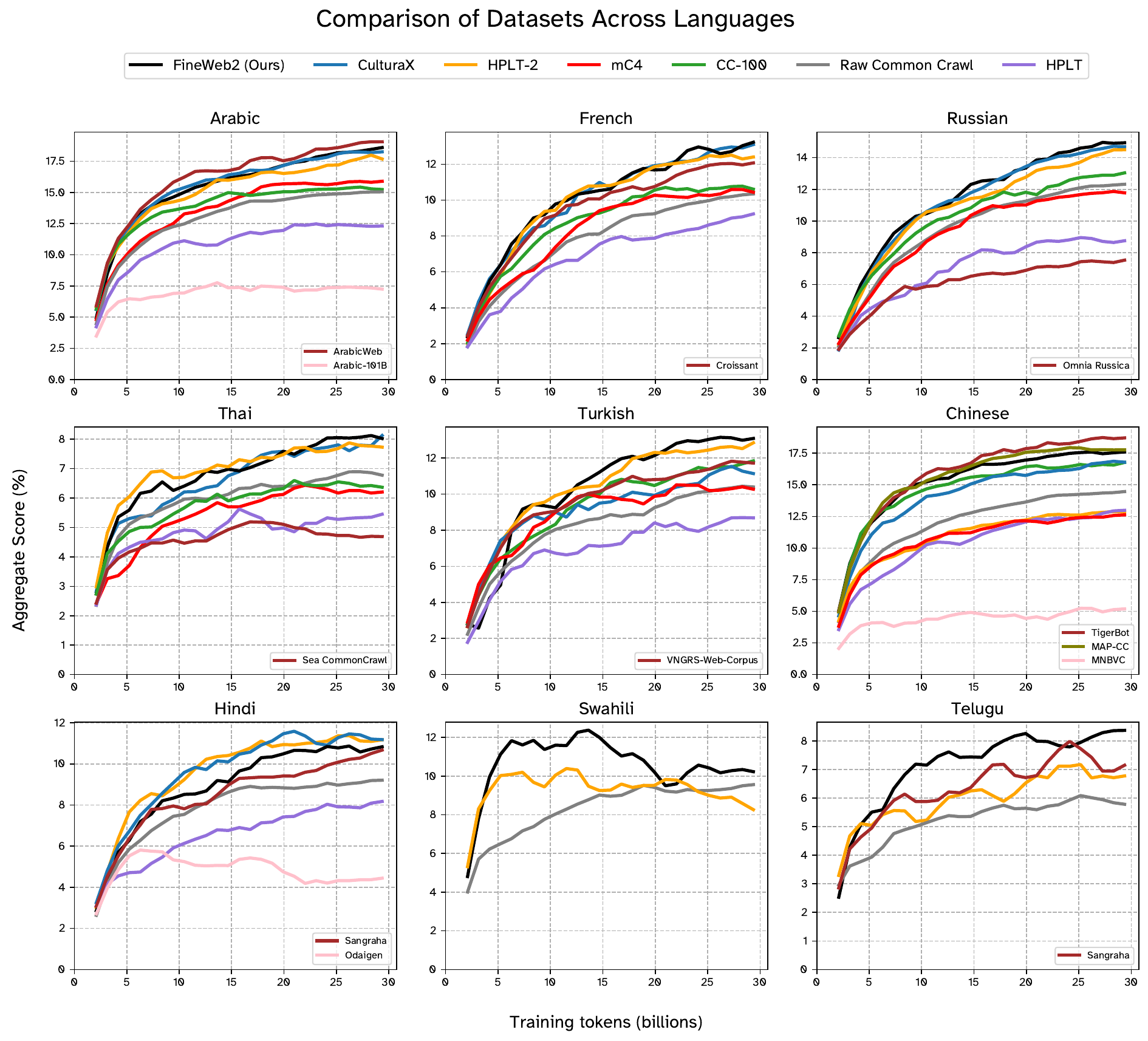}
\caption{\textbf{Per language comparison of FineWeb2} to other multilingual and language-specific datasets. All models were trained for 30 billion tokens. The plots have sliding window smoothing of size 3.}
\label{fig:canary_lang_datasets_comparison}
\end{figure}

\clearpage
\subsection{Dataset comparison on Unseen Languages}
\label{app:unseen}
\subsubsection{List of selected evaluation tasks for unseen languages}

\begin{table}[h]
\begin{center}
\small
\setlength{\tabcolsep}{3pt}
\begin{tabular}{lcc}
\toprule
Task & Metric & Std \\
\midrule
\text{Meta MMLU} \citep{grattafiori2024llama3herdmodels} & Acc (PMI) & 0.0044 \\
\text{Belebele} \citep{belebele} & Acc (Token) & 0.0097 \\
\text{Okapi: Hellaswag} \citep{okapi} & Acc (Token) & 0.0043 \\
\text{X-CODAH} \citep{xcsr} & Acc (Token) & 0.0104 \\
\text{X-CSQA} \citep{xcsr} & Acc (Token) & 0.0040 \\
\text{Mintaka} \citep{mintaka} & F1 & 0.0028 \\
\text{MLQA} \citep{mlqa} & F1 & 0.0192 \\
\text{X-QuAD} \citep{xquad} & F1 & 0.0134 \\
\bottomrule
\end{tabular}
\caption{Selected tasks for German alongside approximate standard deviation of the scores}
\label{tab:task_selection_de}
\end{center}
\end{table}

\begin{table}[h]
\begin{center}
\small
\setlength{\tabcolsep}{3pt}
\begin{tabular}{lccc}
\toprule
Task & Type & Metric & Std \\
\midrule
\text{Okapi: ARC} \citep{okapi} & GK & Acc (PMI) & 0.0093 \\
\text{Indo-MMLU} \citep{indommlu} & GK & Acc (PMI) & 0.0030 \\
\text{Belebele} \citep{belebele} & RC & Acc (Token) & 0.0060 \\
\text{Okapi: Hellaswag} \citep{okapi} & NLU & Acc (Token) & 0.0063 \\
\text{X-COPA} \citep{xcopa} & RES & Acc (Token) & 0.0061 \\
\text{X-Story Cloze} \citep{xstory_cloze} & NLU & Acc (Token) & 0.0053 \\
\text{Tydiqa} \citep{tydiqa} & RC & F1 & 0.0120 \\
\bottomrule
\end{tabular}
\caption{Selected tasks for Indonesian alongside approximate standard deviation of the scores}
\label{tab:task_selection_id}
\end{center}
\end{table}

\begin{table}[h]
\begin{center}
\small
\setlength{\tabcolsep}{3pt}
\begin{tabular}{lccc}
\toprule
Task & Type & Metric & Std \\
\midrule
\text{Okapi: ARC} \citep{okapi} & GK & Acc (PMI) & 0.0119 \\
\text{Meta MMLU} \citep{grattafiori2024llama3herdmodels} & GK & Acc (PMI) & 0.0030 \\
\text{X-CSQA} \citep{xcsr} & RES & Acc (PMI) & 0.0096 \\
\text{Belebele} \citep{belebele} & RC & Acc (Token) & 0.0036 \\
\text{Okapi: Hellaswag} \citep{okapi} & NLU & Acc (Token) & 0.0059 \\
\text{M3Exam} \citep{m3exam} & GK & Acc (Token) & 0.0038 \\
\text{X-CODAH} \citep{xcsr} & RES & Acc (Token) & 0.0203 \\
\text{X-COPA} \citep{xcopa} & RES & Acc (Token) & 0.0059 \\
\text{Mintaka} \citep{mintaka} & GK & F1 & 0.0029 \\
\text{SQuAD-It} \citep{squad_it} & RC & F1 & 0.0155 \\
\bottomrule
\end{tabular}
\caption{Selected tasks for Italian alongside approximate standard deviation of the scores}
\label{tab:task_selection_it}
\end{center}
\end{table}

\begin{table}[h]
\begin{center}
\small
\setlength{\tabcolsep}{3pt}
\begin{tabular}{lccc}
\toprule
Task & Type & Metric & Std \\
\midrule
\text{JMMLU} \citep{jmmlu} & GK & Acc (PMI) & 0.0047 \\
\text{X-CSQA} \citep{xcsr} & RES & Acc (PMI) & 0.0168 \\
\text{Belebele} \citep{belebele} & RC & Acc (Token) & 0.0047 \\
\text{CommonSenseQA} \citep{jglue} & RES & Acc (Token) & 0.0089 \\
\text{X-CODAH} \citep{xcsr} & RES & Acc (Token) & 0.0088 \\
\text{X-Winograd} \citep{xwinograd} & NLU & Acc (Token) & 0.0092 \\
\text{JSQuAD} \citep{jglue} & RC & F1 & 0.0117 \\
\bottomrule
\end{tabular}
\caption{Selected tasks for Japanese alongside approximate standard deviation of the scores}
\label{tab:task_selection_ja}
\end{center}
\end{table}

\begin{table}[h]
\begin{center}
\small
\setlength{\tabcolsep}{3pt}
\begin{tabular}{lccc}
\toprule
Task & Type & Metric & Std \\
\midrule
\text{Okapi: ARC} \citep{okapi} & GK & Acc (PMI) & 0.0045 \\
\text{Okapi: MMLU} \citep{okapi} & GK & Acc (PMI) & 0.0012 \\
\text{X-COPA} \citep{xcopa} & RES & Acc (PMI) & 0.0140 \\
\text{Belebele} \citep{belebele} & RC & Acc (Token) & 0.0148 \\
\text{Okapi: Hellaswag} \citep{okapi} & NLU & Acc (Token) & 0.0099 \\
\text{M3Exam} \citep{m3exam} & GK & Acc (Token) & 0.0080 \\
\text{X-CODAH} \citep{xcsr} & RES & Acc (Token) & 0.0045 \\
\text{X-CSQA} \citep{xcsr} & RES & Acc (Token) & 0.0120 \\
\text{MLQA} \citep{mlqa} & RC & F1 & 0.0118 \\
\text{X-QuAD} \citep{xquad} & RC & F1 & 0.0067 \\
\bottomrule
\end{tabular}
\caption{Selected tasks for Vietnamese alongside approximate standard deviation of the scores}
\label{tab:task_selection_vi}
\end{center}
\end{table}

\break

\clearpage
\subsubsection{Full evaluation results}
\label{app:unseen-results}
% German Table
\begin{table}[h]
\begin{center}
\label{tab:results_de}
\small
\setlength{\tabcolsep}{3pt}
\begin{tabular}{p{2.5cm}|c||cc|cc|cc|cc}
\toprule
 & \multicolumn{1}{c||}{Random} & \multicolumn{2}{c|}{FineWeb2 (ours)} & \multicolumn{2}{c|}{Common Crawl} & \multicolumn{2}{c|}{CulturaX} & \multicolumn{2}{c}{HPLT2} \\
Task & Baseline & Raw & Rescaled & Raw & Rescaled & Raw & Rescaled & Raw & Rescaled \\
\midrule
Belebele (RC) & 25.0 & \textbf{36.6} & \textbf{15.4} & 34.2 & 12.3 & 35.7 & 14.3 & 36.0 & 14.7 \\
Okapi: Hellaswag (NLU) & 25.0 & \textbf{42.5} & \textbf{23.4} & 37.1 & 16.2 & 40.8 & 21.0 & 41.3 & 21.7 \\
X-CODAH (RES) & 25.0 & 39.7 & 19.6 & 39.1 & 18.8 & \textbf{45.0} & \textbf{26.7} & 41.8 & 22.4 \\
X-CSQA (RES) & 20.0 & \textbf{29.1} & \textbf{11.4} & 26.7 & 8.3 & 26.8 & 8.5 & 29.0 & 11.3 \\
Mintaka (GK) & 0.0 & 5.9 & 5.9 & 6.4 & 6.4 & 4.6 & 4.6 & \textbf{7.7} & \textbf{7.7} \\
MLQA (RC) & 0.0 & 28.1 & 28.1 & 26.2 & 26.2 & 28.7 & 28.7 & \textbf{28.9} & \textbf{28.9} \\
X-QuAD (RC) & 0.0 & \textbf{26.2} & \textbf{26.2} & 24.3 & 24.3 & 23.7 & 23.7 & 24.3 & 24.3 \\
Meta MMLU (GK) & 25.0 & 29.5 & 6.0 & 27.9 & 3.8 & 29.0 & 5.3 & \textbf{30.0} & \textbf{6.7} \\
\midrule
GK tasks & - & \multicolumn{2}{c|}{6.0} & \multicolumn{2}{c|}{5.1} & \multicolumn{2}{c|}{5.0} & \multicolumn{2}{c}{\textbf{7.2}} \\
RC tasks & - & \multicolumn{2}{c|}{\textbf{23.2}} & \multicolumn{2}{c|}{20.9} & \multicolumn{2}{c|}{22.2} & \multicolumn{2}{c}{22.6} \\
RES tasks & - & \multicolumn{2}{c|}{15.5} & \multicolumn{2}{c|}{13.6} & \multicolumn{2}{c|}{\textbf{17.6}} & \multicolumn{2}{c}{16.8} \\
NLU tasks & - & \multicolumn{2}{c|}{\textbf{23.4}} & \multicolumn{2}{c|}{16.2} & \multicolumn{2}{c|}{21.0} & \multicolumn{2}{c}{21.7} \\
\midrule
Aggregate Score & - & \multicolumn{2}{c|}{17.0} & \multicolumn{2}{c|}{13.9} & \multicolumn{2}{c|}{16.4} & \multicolumn{2}{c}{\textbf{17.1}} \\
\bottomrule
\end{tabular}
\caption{German Results}
\end{center}
\end{table}

% Indonesian Table
\begin{table}[h]
\begin{center}
\label{tab:results_id}
\small
\setlength{\tabcolsep}{3pt}
\begin{tabular}{p{2.5cm}|c||cc|cc|cc|cc}
\toprule
 & \multicolumn{1}{c||}{Random} & \multicolumn{2}{c|}{FineWeb2 (ours)} & \multicolumn{2}{c|}{Common Crawl} & \multicolumn{2}{c|}{CulturaX} & \multicolumn{2}{c}{HPLT2} \\
Task & Baseline & Raw & Rescaled & Raw & Rescaled & Raw & Rescaled & Raw & Rescaled \\
\midrule
Okapi: ARC (GK) & 25.0 & 30.8 & 7.8 & 29.1 & 5.4 & 30.5 & 7.4 & \textbf{33.7} & \textbf{11.6} \\
Belebele (RC) & 25.0 & 31.8 & 9.1 & 32.0 & 9.3 & \textbf{32.3} & \textbf{9.7} & 32.1 & 9.5 \\
Okapi: Hellaswag (NLU) & 25.0 & 41.4 & 21.9 & 38.6 & 18.1 & 41.8 & 22.4 & \textbf{42.7} & \textbf{23.6} \\
X-COPA (RES) & 50.0 & 63.3 & 26.5 & 60.9 & 21.7 & 65.9 & 31.9 & \textbf{66.2} & \textbf{32.4} \\
X-Story Cloze (NLU) & 50.0 & \textbf{66.0} & \textbf{32.1} & 63.6 & 27.1 & 63.9 & 27.9 & 65.7 & 31.5 \\
Tydiqa (RC) & 0.0 & 33.6 & 33.6 & \textbf{34.6} & \textbf{34.6} & 29.0 & 29.0 & 32.3 & 32.3 \\
Indo-MMLU (GK) & 25.0 & 28.9 & 5.2 & 28.7 & 4.9 & 28.0 & 4.0 & \textbf{29.6} & \textbf{6.1} \\
\midrule
GK tasks & - & \multicolumn{2}{c|}{6.5} & \multicolumn{2}{c|}{5.1} & \multicolumn{2}{c|}{5.7} & \multicolumn{2}{c}{\textbf{8.9}} \\
RC tasks & - & \multicolumn{2}{c|}{21.4} & \multicolumn{2}{c|}{\textbf{21.9}} & \multicolumn{2}{c|}{19.4} & \multicolumn{2}{c}{20.9} \\
RES tasks & - & \multicolumn{2}{c|}{26.5} & \multicolumn{2}{c|}{21.7} & \multicolumn{2}{c|}{31.9} & \multicolumn{2}{c}{\textbf{32.4}} \\
NLU tasks & - & \multicolumn{2}{c|}{27.0} & \multicolumn{2}{c|}{22.6} & \multicolumn{2}{c|}{25.2} & \multicolumn{2}{c}{\textbf{27.6}} \\
\midrule
Aggregate Score & - & \multicolumn{2}{c|}{20.3} & \multicolumn{2}{c|}{17.9} & \multicolumn{2}{c|}{20.5} & \multicolumn{2}{c}{\textbf{22.4}} \\
\bottomrule
\end{tabular}
\caption{Indonesian Results}
\end{center}
\end{table}

% Italian Table
\begin{table}[h]
\begin{center}
\label{tab:results_it}
\small
\setlength{\tabcolsep}{3pt}
\begin{tabular}{p{2.5cm}|c||cc|cc|cc|cc}
\toprule
 & \multicolumn{1}{c||}{Random} & \multicolumn{2}{c|}{FineWeb2 (ours)} & \multicolumn{2}{c|}{Common Crawl} & \multicolumn{2}{c|}{CulturaX} & \multicolumn{2}{c}{HPLT2} \\
Task & Baseline & Raw & Rescaled & Raw & Rescaled & Raw & Rescaled & Raw & Rescaled \\
\midrule
Okapi: ARC (GK) & 25.0 & \textbf{32.4} & \textbf{9.9} & 28.7 & 4.9 & 30.0 & 6.6 & 30.7 & 7.6 \\
Belebele (RC) & 25.0 & \textbf{31.9} & \textbf{9.2} & 28.7 & 5.0 & 30.5 & 7.4 & 30.4 & 7.2 \\
Okapi: Hellaswag (NLU) & 25.0 & \textbf{45.4} & \textbf{27.2} & 38.5 & 18.0 & 43.6 & 24.8 & 44.4 & 25.8 \\
M3Exam (GK) & 33.8 & 39.1 & 8.0 & 38.3 & 6.8 & \textbf{40.0} & \textbf{9.5} & 38.6 & 7.3 \\
X-CODAH (RES) & 25.0 & \textbf{39.3} & \textbf{19.1} & 38.7 & 18.2 & 38.0 & 17.3 & 38.7 & 18.2 \\
X-CSQA (RES) & 20.0 & 37.5 & 21.9 & 32.8 & 16.0 & \textbf{37.6} & \textbf{21.9} & 36.1 & 20.2 \\
X-COPA (RES) & 50.0 & 64.8 & 29.6 & 61.7 & 23.3 & 63.0 & 26.0 & \textbf{65.2} & \textbf{30.4} \\
Mintaka (GK) & 0.0 & 10.4 & 10.4 & 7.9 & 7.9 & 9.8 & 9.8 & \textbf{10.6} & \textbf{10.6} \\
SQuAD-It (RC) & 0.0 & 20.3 & 20.3 & 18.2 & 18.2 & \textbf{22.2} & \textbf{22.2} & 21.8 & 21.8 \\
Meta MMLU (GK) & 25.0 & \textbf{30.1} & \textbf{6.7} & 29.0 & 5.3 & 29.1 & 5.5 & 29.5 & 5.9 \\
\midrule
GK tasks & - & \multicolumn{2}{c|}{\textbf{8.8}} & \multicolumn{2}{c|}{6.2} & \multicolumn{2}{c|}{7.8} & \multicolumn{2}{c}{7.9} \\
RC tasks & - & \multicolumn{2}{c|}{14.7} & \multicolumn{2}{c|}{11.6} & \multicolumn{2}{c|}{\textbf{14.8}} & \multicolumn{2}{c}{14.5} \\
RES tasks & - & \multicolumn{2}{c|}{\textbf{23.5}} & \multicolumn{2}{c|}{19.2} & \multicolumn{2}{c|}{21.8} & \multicolumn{2}{c}{22.9} \\
NLU tasks & - & \multicolumn{2}{c|}{\textbf{27.2}} & \multicolumn{2}{c|}{18.0} & \multicolumn{2}{c|}{24.8} & \multicolumn{2}{c}{25.8} \\
\midrule
Aggregate Score & - & \multicolumn{2}{c|}{\textbf{18.6}} & \multicolumn{2}{c|}{13.7} & \multicolumn{2}{c|}{17.3} & \multicolumn{2}{c}{17.8} \\
\bottomrule
\end{tabular}
\caption{Italian Results}
\end{center}
\end{table}

% Japanese Table
\begin{table}[h]
\begin{center}
\label{tab:results_ja}
\small
\setlength{\tabcolsep}{3pt}
\begin{tabular}{p{2.5cm}|c||cc|cc|cc|cc}
\toprule
 & \multicolumn{1}{c||}{Random} & \multicolumn{2}{c|}{FineWeb2 (ours)} & \multicolumn{2}{c|}{Common Crawl} & \multicolumn{2}{c|}{CulturaX} & \multicolumn{2}{c}{HPLT2} \\
Task & Baseline & Raw & Rescaled & Raw & Rescaled & Raw & Rescaled & Raw & Rescaled \\
\midrule
Belebele (RC) & 25.0 & \textbf{32.5} & \textbf{10.0} & 31.7 & 8.9 & 30.3 & 7.1 & 29.3 & 5.8 \\
CommonSenseQA (RES) & 20.0 & \textbf{67.5} & \textbf{59.4} & 60.9 & 51.2 & 63.5 & 54.4 & 50.3 & 37.8 \\
X-CODAH (RES) & 25.0 & 37.7 & 16.9 & 37.7 & 16.9 & \textbf{38.7} & \textbf{18.2} & 37.4 & 16.6 \\
X-CSQA (RES) & 20.0 & 36.4 & 20.5 & 36.4 & 20.5 & \textbf{37.2} & \textbf{21.5} & 31.0 & 13.7 \\
X-Winograd (NLU) & 50.0 & \textbf{60.3} & \textbf{20.6} & 54.4 & 8.9 & 57.7 & 15.4 & 59.0 & 18.0 \\
JSQuAD (RC) & 0.0 & \textbf{40.5} & \textbf{40.5} & 33.1 & 33.1 & 28.5 & 28.5 & 11.7 & 11.7 \\
JMMLU (GK) & 25.0 & \textbf{31.7} & \textbf{9.0} & 28.9 & 5.1 & 30.7 & 7.5 & 28.7 & 4.9 \\
\midrule
GK tasks & - & \multicolumn{2}{c|}{\textbf{9.0}} & \multicolumn{2}{c|}{5.1} & \multicolumn{2}{c|}{7.5} & \multicolumn{2}{c}{4.9} \\
RC tasks & - & \multicolumn{2}{c|}{\textbf{25.3}} & \multicolumn{2}{c|}{21.0} & \multicolumn{2}{c|}{17.8} & \multicolumn{2}{c}{8.7} \\
RES tasks & - & \multicolumn{2}{c|}{\textbf{32.2}} & \multicolumn{2}{c|}{29.5} & \multicolumn{2}{c|}{31.4} & \multicolumn{2}{c}{22.7} \\
NLU tasks & - & \multicolumn{2}{c|}{\textbf{20.6}} & \multicolumn{2}{c|}{8.9} & \multicolumn{2}{c|}{15.4} & \multicolumn{2}{c}{18.0} \\
\midrule
Aggregate Score & - & \multicolumn{2}{c|}{\textbf{21.8}} & \multicolumn{2}{c|}{16.1} & \multicolumn{2}{c|}{18.0} & \multicolumn{2}{c}{13.6} \\
\bottomrule
\end{tabular}
\caption{Japanese Results}
\end{center}
\end{table}

% Vietnamese Table
\begin{table}[h]
\begin{center}
\label{tab:results_vi}
\small
\setlength{\tabcolsep}{3pt}
\begin{tabular}{p{2.5cm}|c||cc|cc|cc|cc}
\toprule
 & \multicolumn{1}{c||}{Random} & \multicolumn{2}{c|}{FineWeb2 (ours)} & \multicolumn{2}{c|}{Common Crawl} & \multicolumn{2}{c|}{CulturaX} & \multicolumn{2}{c}{HPLT2} \\
Task & Baseline & Raw & Rescaled & Raw & Rescaled & Raw & Rescaled & Raw & Rescaled \\
\midrule
Okapi: ARC (GK) & 25.0 & \textbf{31.3} & \textbf{8.4} & 27.2 & 2.9 & 30.8 & 7.7 & 31.1 & 8.1 \\
Belebele (RC) & 25.0 & 33.0 & 10.6 & 32.6 & 10.2 & 33.1 & 10.8 & \textbf{34.1} & \textbf{12.1} \\
Okapi: Hellaswag (NLU) & 25.0 & \textbf{48.7} & \textbf{31.6} & 43.2 & 24.2 & 46.6 & 28.8 & 44.5 & 26.0 \\
M3Exam (GK) & 25.2 & 35.2 & 13.3 & 36.7 & 15.4 & 38.0 & 17.0 & \textbf{39.1} & \textbf{18.6} \\
X-CODAH (RES) & 25.0 & \textbf{40.3} & \textbf{20.4} & 35.6 & 14.1 & 38.2 & 17.6 & 38.4 & 17.9 \\
X-CSQA (RES) & 20.0 & 29.6 & 12.0 & 28.5 & 10.6 & \textbf{29.8} & \textbf{12.3} & 29.7 & 12.2 \\
X-COPA (RES) & 50.0 & \textbf{75.7} & \textbf{51.3} & 69.7 & 39.5 & 64.6 & 29.2 & 70.5 & 40.9 \\
MLQA (RC) & 0.0 & 19.4 & 19.4 & 18.6 & 18.6 & \textbf{23.4} & \textbf{23.4} & 22.3 & 22.3 \\
X-QuAD (RC) & 0.0 & 17.3 & 17.3 & 16.9 & 16.9 & \textbf{21.3} & \textbf{21.3} & 21.2 & 21.2 \\
Okapi: MMLU (GK) & 25.0 & \textbf{29.4} & \textbf{5.8} & 28.5 & 4.7 & 28.1 & 4.1 & 28.8 & 5.0 \\
\midrule
GK tasks & - & \multicolumn{2}{c|}{9.2} & \multicolumn{2}{c|}{7.6} & \multicolumn{2}{c|}{9.6} & \multicolumn{2}{c}{\textbf{10.6}} \\
RC tasks & - & \multicolumn{2}{c|}{15.8} & \multicolumn{2}{c|}{15.3} & \multicolumn{2}{c|}{18.5} & \multicolumn{2}{c}{\textbf{18.5}} \\
RES tasks & - & \multicolumn{2}{c|}{\textbf{27.9}} & \multicolumn{2}{c|}{21.4} & \multicolumn{2}{c|}{19.7} & \multicolumn{2}{c}{23.7} \\
NLU tasks & - & \multicolumn{2}{c|}{\textbf{31.6}} & \multicolumn{2}{c|}{24.2} & \multicolumn{2}{c|}{28.8} & \multicolumn{2}{c}{26.0} \\
\midrule
Aggregate Score & - & \multicolumn{2}{c|}{\textbf{21.1}} & \multicolumn{2}{c|}{17.1} & \multicolumn{2}{c|}{19.2} & \multicolumn{2}{c}{19.7} \\
\bottomrule
\end{tabular}
\caption{Vietnamese Results}
\end{center}
\end{table}

\clearpage
\subsection{FineWeb2 language composition}\label{app:fineweb2-composition}
\begin{figure}[h]
\centering
\includegraphics[width=\textwidth]{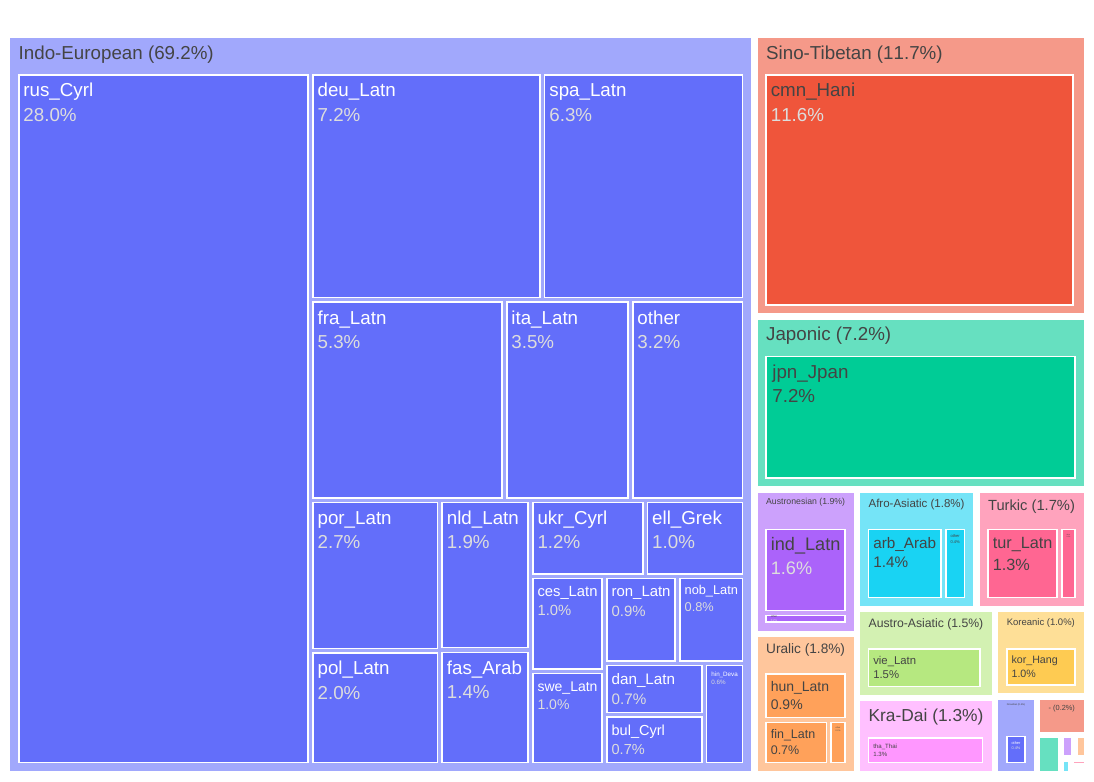}
\caption{\textbf{Language composition of FineWeb2} Distribution of languages in the final FineWeb2 dataset. Percentages refer to total utf-8 bytes of each language or language family.}
\label{fig:canary_lang_datasets_comparison}
\end{figure}
\clearpage

\setlength{\tabcolsep}{3pt}
\begin{longtable}{p{1cm} l p{3cm} p{2cm} lll}
  \caption{FineWeb2 80 largest language stats}\label{tab:fw2-per-language}\\
  
  % Header for first page
  \toprule
ISO 639-3 & Script & Name & Language Family & Words & Documents & Disk size \\
  \midrule
  \endfirsthead
  
  % Header for subsequent pages
  \multicolumn{7}{c}{Table \thetable\ -- Continued from previous page} \\
  \toprule
ISO 639-3 & Script & Name & Language Family & Words & Documents & Disk size \\
  \midrule
  \endhead
  
  % Footer for all pages except last
  \midrule
  % \multicolumn{7}{r}{Continued on next page} \\
  \endfoot
  
  % Footer for last page
  \bottomrule
  \endlastfoot

rus & Cyrl & Russian & Indo-European & 588,579,493,780 & 699,083,579 & 5.82TB \\
cmn & Hani & Mandarin Chinese & Sino-Tibetan & 543,543,038,750 & 636,058,984 & 2.42TB \\
deu & Latn & German & Indo-European & 262,271,052,199 & 495,964,485 & 1.51TB \\
jpn & Jpan & Japanese & Japonic & 331,144,301,801 & 400,138,563 & 1.50TB \\
spa & Latn & Spanish & Indo-European & 261,523,749,595 & 441,287,261 & 1.32TB \\
fra & Latn & French & Indo-European & 220,662,584,640 & 360,058,973 & 1.11TB \\
ita & Latn & Italian & Indo-European & 139,116,026,491 & 238,984,437 & 739.24GB \\
por & Latn & Portuguese & Indo-European & 109,536,087,117 & 199,737,979 & 569.24GB \\
pol & Latn & Polish & Indo-European & 73,119,437,217 & 151,966,724 & 432.01GB \\
nld & Latn & Dutch & Indo-European & 74,634,633,118 & 147,301,270 & 397.51GB \\
ind & Latn & Indonesian & Austronesian & 60,264,322,142 & 100,238,529 & 348.65GB \\
vie & Latn & Vietnamese & Austro-Asiatic & 50,886,874,358 & 61,064,248 & 319.83GB \\
fas & Arab & Persian & Indo-European & 39,705,799,658 & 58,843,652 & 304.62GB \\
arb & Arab & Standard Arabic & Afro-Asiatic & 32,812,858,120 & 61,977,525 & 293.59GB \\
tur & Latn & Turkish & Turkic & 41,933,799,420 & 95,129,129 & 284.52GB \\
tha & Thai & Thai & Kra-Dai & 24,662,748,945 & 35,897,202 & 278.68GB \\
ukr & Cyrl & Ukrainian & Indo-European & 25,586,457,655 & 53,101,726 & 254.86GB \\
ell & Grek & Modern Greek (1453-) & Indo-European & 22,827,957,288 & 47,421,073 & 222.05GB \\
kor & Hang & Korean & Koreanic & 48,613,120,582 & 60,874,355 & 213.43GB \\
ces & Latn & Czech & Indo-European & 35,479,428,809 & 66,067,904 & 206.33GB \\
swe & Latn & Swedish & Indo-European & 35,745,969,364 & 59,485,306 & 202.96GB \\
hun & Latn & Hungarian & Uralic & 30,919,839,164 & 49,935,986 & 199.69GB \\
ron & Latn & Romanian & Indo-European & 35,017,893,659 & 58,303,671 & 186.19GB \\
nob & Latn & Norwegian Bokmål & Indo-European & 32,008,904,934 & 38,144,343 & 172.05GB \\
dan & Latn & Danish & Indo-European & 28,055,948,840 & 45,391,655 & 150.72GB \\
bul & Cyrl & Bulgarian & Indo-European & 16,074,326,712 & 25,994,731 & 145.75GB \\
fin & Latn & Finnish & Uralic & 20,343,096,672 & 36,710,816 & 143.03GB \\
hin & Deva & Hindi & Indo-European & 11,173,681,651 & 22,095,985 & 120.98GB \\
ben & Beng & Bengali & Indo-European & 6,153,579,265 & 15,185,742 & 87.04GB \\
slk & Latn & Slovak & Indo-European & 14,808,010,769 & 29,991,521 & 85.43GB \\
heb & Hebr & Hebrew & Afro-Asiatic & 8,462,976,117 & 14,491,748 & 68.71GB \\
lit & Latn & Lithuanian & Indo-European & 9,132,828,961 & 13,471,965 & 56.50GB \\
bos & Latn & Bosnian & Indo-European & 9,086,837,979 & 21,243,255 & 49.18GB \\
slv & Latn & Slovenian & Indo-European & 7,688,373,264 & 12,059,130 & 41.80GB \\
ekk & Latn & Standard Estonian & Uralic & 6,564,292,000 & 10,218,587 & 40.82GB \\
cat & Latn & Catalan & Indo-European & 8,348,091,726 & 17,136,414 & 40.35GB \\
tam & Taml & Tamil & Dravidian & 1,937,150,898 & 5,528,854 & 36.97GB \\
hrv & Latn & Croatian & Indo-European & 6,609,299,440 & 6,195,824 & 35.91GB \\
lvs & Latn & Standard Latvian & Indo-European & 5,371,151,279 & 8,030,316 & 33.36GB \\
zsm & Latn & Standard Malay & Austronesian & 5,648,387,840 & 9,421,248 & 31.94GB \\
azj & Latn & North Azerbaijani & Turkic & 3,894,255,826 & 7,291,231 & 26.90GB \\
srp & Cyrl & Serbian & Indo-European & 2,858,500,314 & 4,146,124 & 26.87GB \\
kat & Geor & Georgian & Kartvelian & 1,439,572,993 & 3,706,659 & 25.23GB \\
npi & Deva & Nepali (individual language) & Indo-European & 1,642,856,349 & 4,888,163 & 25.13GB \\
mar & Deva & Marathi & Indo-European & 1,541,225,070 & 3,912,702 & 22.57GB \\
mal & Mlym & Malayalam & Dravidian & 1,054,187,581 & 3,322,526 & 22.27GB \\
kaz & Cyrl & Kazakh & Turkic & 1,876,843,453 & 3,344,366 & 20.67GB \\
urd & Arab & Urdu & Indo-European & 2,733,266,493 & 4,809,542 & 19.93GB \\
als & Latn & Tosk Albanian & Indo-European & 3,454,387,059 & 8,597,826 & 18.18GB \\
mkd & Cyrl & Macedonian & Indo-European & 1,611,392,841 & 4,150,902 & 14.99GB \\
tel & Telu & Telugu & Dravidian & 891,002,487 & 1,964,395 & 14.42GB \\
kan & Knda & Kannada & Dravidian & 748,850,327 & 2,390,982 & 12.91GB \\
mya & Mymr & Burmese & Sino-Tibetan & 854,400,671 & 1,558,304 & 12.35GB \\
guj & Gujr & Gujarati & Indo-European & 934,124,052 & 2,127,094 & 11.71GB \\
bel & Cyrl & Belarusian & Indo-European & 1,166,541,148 & 2,100,873 & 11.47GB \\
isl & Latn & Icelandic & Indo-European & 1,696,354,360 & 3,014,429 & 10.27GB \\
khm & Khmr & Khmer & Austro-Asiatic & 667,495,692 & 1,586,460 & 8.70GB \\
khk & Cyrl & Halh Mongolian & Mongolic & 824,211,882 & 1,622,882 & 8.52GB \\
fil & Latn & Filipino & Austronesian & 1,636,238,017 & 2,349,050 & 8.13GB \\
ary & Arab & Moroccan Arabic & Afro-Asiatic & 843,523,994 & 2,365,405 & 7.74GB \\
afr & Latn & Afrikaans & Indo-European & 1,598,352,868 & 1,992,040 & 7.69GB \\
hye & Armn & Armenian & Indo-European & 634,273,060 & 1,757,415 & 7.17GB \\
sin & Sinh & Sinhala & Indo-European & 512,453,069 & 1,185,323 & 7.05GB \\
glg & Latn & Galician & Indo-European & 1,236,233,473 & 2,522,814 & 6.47GB \\
uzn & Cyrl & Northern Uzbek & Turkic & 544,866,919 & 1,357,811 & 6.12GB \\
pan & Guru & Panjabi & Indo-European & 522,788,467 & 944,160 & 5.64GB \\
ory & Orya & Odia & Indo-European & 333,760,951 & 1,298,188 & 4.92GB \\
uzn & Latn & Northern Uzbek & Turkic & 687,002,994 & 1,233,463 & 4.45GB \\
kir & Cyrl & Kirghiz & Turkic & 397,449,282 & 1,069,582 & 4.36GB \\
eus & Latn & Basque & Language isolate & 711,939,889 & 1,569,434 & 4.30GB \\
lat & Latn & Latin & Indo-European & 714,764,848 & 1,473,541 & 3.86GB \\
tgk & Cyrl & Tajik & Indo-European & 396,209,383 & 688,384 & 3.75GB \\
gmh & Latn & Middle High German (ca. 1050-1500) & Indo-European & 506,396,917 & 84,495 & 3.41GB \\
swh & Latn & Swahili (individual language) & Niger-Congo & 569,542,024 & 1,206,300 & 3.08GB \\
arz & Arab & Egyptian Arabic & Afro-Asiatic & 345,040,810 & 853,290 & 2.92GB \\
nno & Latn & Norwegian Nynorsk & Indo-European & 522,740,774 & 1,214,870 & 2.68GB \\
cym & Latn & Welsh & Indo-European & 523,226,616 & 831,878 & 2.50GB \\
amh & Ethi & Amharic & Afro-Asiatic & 239,936,286 & 428,373 & 2.49GB \\
pbt & Arab & Southern Pashto & Indo-European & 337,138,269 & 639,983 & 2.41GB \\
ckb & Arab & Central Kurdish & Indo-European & 236,342,609 & 554,993 & 2.39GB \\

\ldots\footnote{Full list available at \url{https://github.com/huggingface/fineweb-2/blob/main/fineweb2-language-distribution.csv}} & & & & & & \\
\midrule
\textbf{Total} & & & & \textbf{3,339,271,691,958} & \textbf{5,018,505,566} & \textbf{20.78TB} \\
\end{longtable}

\clearpage
\subsection{Bible and Wikipedia content}\label{app:lower-resource-wiki-bible}
\begin{figure}[h]
\centering
\includegraphics[width=\textwidth]{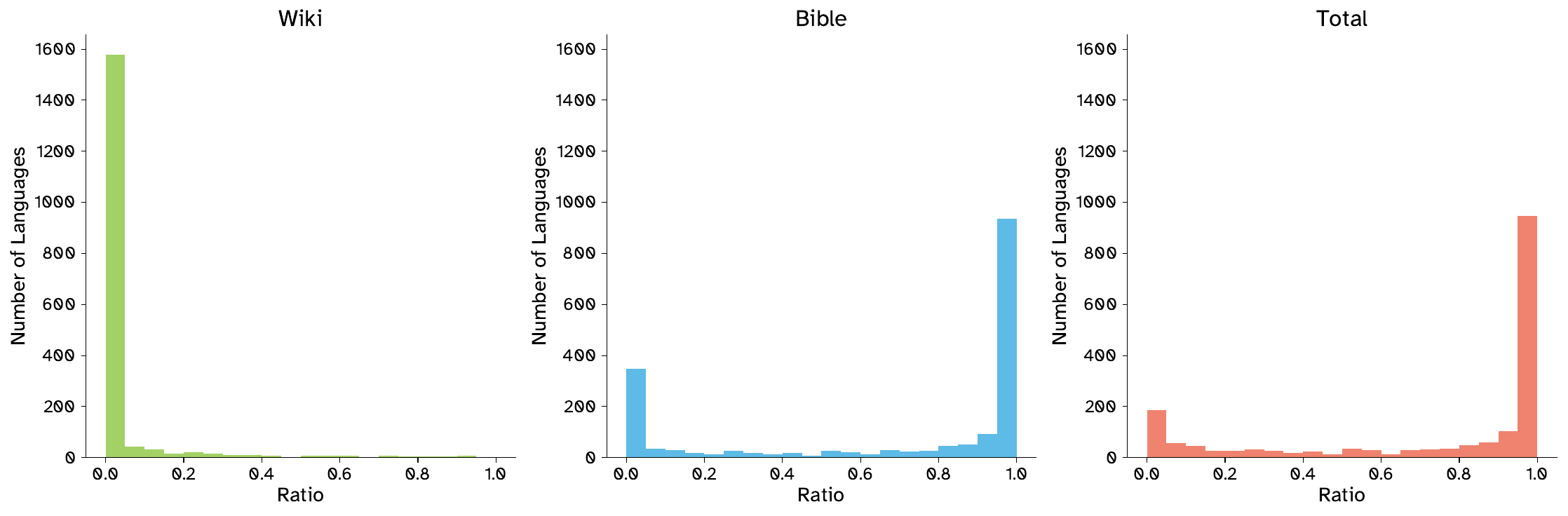}
\caption{\textbf{Ratio of Wikipedia and Bible content per language} Most languages have a small fraction of their content originating from Wikipedia (with some exceptions). Bible content, on the other hand, is a big part of the corpora of many lower-resource languages.}
\label{fig:bible_wiki_ratios}
\end{figure}

For each language low resource language, we first compiled the distribution of documents by domain name. We then averaged the frequency of each domain across all languages, to find specific domains that were a common source of data for different languages (which from manual inspection was the case for specific Bible websites and Wikipedia). We manually labeled the top domains that belonged to Bible or Wikipedia websites (\cref{tab:bible-wiki-domains}), and then measured the fraction of each language corpora that belonged to these domains. Out of 1868 language-script pairs in the final dataset, 70\% (1320 of them) have more than half their documents from Bible- or Wikipedia-related domains. This is mostly driven by Bible content, as can be seen in~\cref{fig:bible_wiki_ratios}.

\begin{table}[h]
\begin{center}
\small
\setlength{\tabcolsep}{3pt}
\begin{tabular}{ll}
\toprule
\textbf{Bible Domains} & \textbf{Wiki Domains} \\
\midrule
\text{ebible.org} & \text{wikipedia.org} \\
\text{bible.is} & \text{wikimedia.org} \\
\text{jw.org} & \text{wikisource.org} \\
\text{stepbible.org} & \text{wiktionary.org} \\
\text{bibles.org} & \\
\text{bible.com} & \\
\text{breakeveryyoke.com} & \\
\text{png.bible} & \\
\text{americanbible.org} & \\
\text{pngscriptures.org} & \\
\text{globalrecordings.net} & \\
\text{gospelgo.com} & \\
\text{httlvn.org} & \\
\text{biblegateway.com} & \\
\text{jesusforafrica.net} & \\
\text{bible.com.au} & \\
\text{pacificbibles.org} & \\
\text{scriptureearth.org} & \\
\text{divinerevelations.info} & \\
\text{beblia.com} & \\
\text{aboriginalbibles.org.au} & \\
\text{eevangelize.com} & \\
\text{biblica.com} & \\
\text{e-alkitab.org} & \\
\text{alkitab.pw} & \\
\text{amazinggracebibleinstitute.com} & \\
\text{bibleforchildren.org} & \\
\text{aionianbible.org} & \\
\text{cyber.bible} & \\
\text{biblehub.com} & \\
\text{myanmarbs.org} & \\
\text{baebol.org} & \\
\text{christianchildmultilingualbibleverse.wordpress.com} & \\
\text{femissionaria.blogspot.com} & \\
\text{biblics.com} & \\
\text{churchofjesuschrist.org} & \\
\text{biblesa.co.za} & \\
\text{bible-tools.org} & \\
\text{torresstraitbibles.org.au} & \\
\bottomrule
\end{tabular}
\caption{List of Bible-related and Wiki-related domains}
\label{tab:bible-wiki-domains}
\end{center}
\end{table}

\clearpage
\subsection{Train-Test Split}
Our dataset release is split into a train and test set, per language.
The test set should not be used for training but instead can help research questions such as on memorization or data attribution.
The test set is obtained as a random subset (by a hash function applied on the document content), and contains $\min\{1\%, 100k\}$ of the documents per language pre-filtering, with a reduction in size when these documents are filtered with the same process as the train set. It is only provided for languages of sufficient size.

\end{document}